\documentclass[runningheads]{llncs}

\usepackage{eccv}

\usepackage{eccvabbrv}

\usepackage{graphicx}
\usepackage{booktabs}

\usepackage[accsupp]{axessibility}  

\usepackage{hyperref}

\usepackage{orcidlink}

\usepackage{makecell}
\usepackage[table]{xcolor}
\usepackage{amssymb}
\usepackage{pifont}
\usepackage{utfsym}
\usepackage{multirow}
\usepackage{algorithm}
\usepackage{algorithmic}

\usepackage{xcolor}
\definecolor{p}{RGB}{178,34,34}
\definecolor{d}{RGB}{65, 105, 225}
\usepackage{diagbox}
\usepackage{comment}

\begin{document}

\title{Spectral Defense Against Resource-Targeting Attack in 3D Gaussian Splatting} 

\titlerunning{Spectral Defense}


\author{Yang Chen\inst{1} \and
Yi Yu\inst{1} \and
Jiaming He\inst{2} \and
Yueqi Duan\inst{3} \and \\ 
Zheng Zhu\inst{4} \and
Yap-Peng Tan\inst{5,1}}

\authorrunning{Y. Chen, Y. Yu et al.}


\institute{$^{1}$Nanyang Technological University  \mbox{~} $^{2}$ UESTC \mbox{~} $^{3}$ Tsinghua University \\
$^{4}$ GigaAI  \mbox{~} $^{5}$ VinUniversity}

\maketitle

\begin{abstract}
  Recent advances in 3D Gaussian Splatting (3DGS) deliver high-quality rendering, yet the Gaussian representation exposes a new attack surface, the resource-targeting attack. This attack poisons training images, excessively inducing  Gaussian growth to cause resource exhaustion. Although efficiency-oriented methods such as smoothing, thresholding, and pruning have been explored, these spatial-domain strategies operate on visible structures but overlook how stealthy perturbations distort the underlying spectral behaviors of training data. As a result, poisoned inputs introduce abnormal high-frequency amplifications that mislead 3DGS into interpreting noisy patterns as detailed structures, ultimately causing unstable Gaussian overgrowth and degraded scene fidelity. To address this, we propose \textbf{Spectral Defense} in Gaussian and image fields. We first design a 3D frequency filter to selectively prune Gaussians exhibiting abnormally high frequencies. Since natural scenes also contain legitimate high-frequency structures, directly suppressing high frequencies is insufficient, and we further develop a 2D spectral regularization on renderings, distinguishing naturally isotropic frequencies while penalizing anisotropic angular energy to constrain noisy patterns. Experiments show that our defense builds robust, accurate, and secure 3DGS, suppressing overgrowth by up to $5.92\times$, reducing memory by up to $3.66\times$, and improving speed by up to $4.34\times$ under attacks.
  
\end{abstract}

\section{Introduction}
\label{sec:intro}

\begin{figure}[t]
  \centering
   \includegraphics[width=1.0\linewidth]{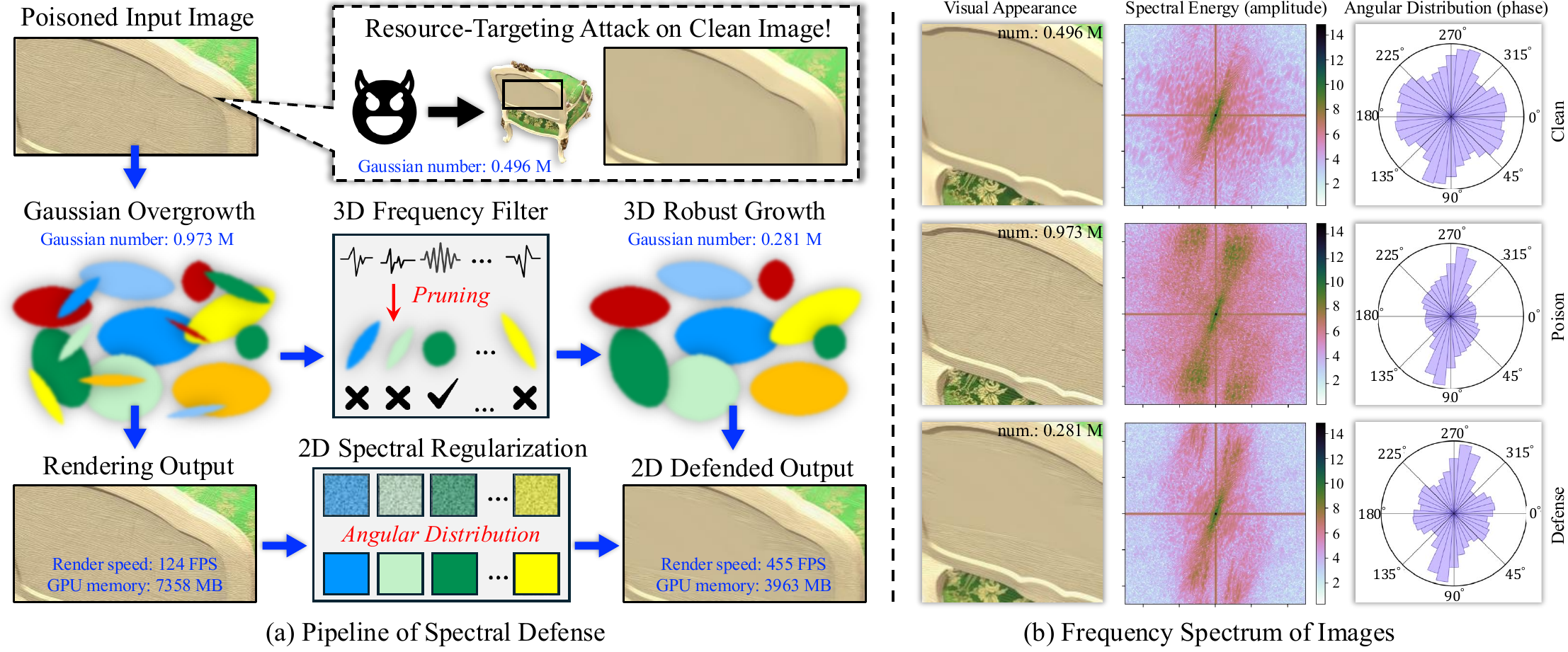}
   \vspace{-4mm}
   \caption{Spectral defense against attack in 3DGS. (a) A resource-targeting attack poisons input images to trigger excessive Gaussian growth, leading to redundant splats and degrading renderings. Our defense operates jointly in 3D and 2D domains, where 3D frequency filter prunes Gaussians with abnormally high-frequency responses to suppress attack-induced overgrowth, and 2D spectral regularization constrains the angular distribution of energy to reduce anisotropic noise. (b) Frequency spectrum reveals spectral energy and angular distribution to distinguish frequency components of images.}
   \label{fig:teaser}
   \vspace{-2mm}
\end{figure}

3DGS~\cite{kerbl20233d} has become a dominant paradigm for scene reconstruction and novel view synthesis, achieving photo-realistic quality with high rendering speed. In 3DGS, a scene is represented by a set of 3D Gaussian primitives, dynamically adjusting to match the complexity of the scene during training, which is optimized from input images~\cite{bao20253d, xu2025depthsplat, kong2025rogsplat, lei2025gaussnav, zhang2025quadratic}. This adaptive mechanism, while key to efficiency and detail of 3DGS, also introduces a new attack surface. Recent work~\cite{lu2024poison} exposes a weakness of Gaussian representations: a data poisoning threat model can target the 3DGS victim model by exploiting its flexibility through a resource-targeting attack~\cite{shumailov2021sponge, chen2023dark, cina2025energy}. This attack only manipulates the training inputs, where poisoned images contain subtle noisy patterns that mislead the 3DGS optimization into excessive Gaussian growth. Thus, the overgrowth perturbs 3DGS with extreme computational consumption and rendering slowdowns.

To defend against the attack, a straightforward idea~\cite{lu2024poison} is smoothing poisoned images or imposing universal Gaussian thresholds. These strategies reveal drawbacks: image smoothing destroys essential detailed structures, while universal Gaussian thresholds fail to generalize across scenes, thereby being overly strict for some scenes and insufficient for others. Inspired by efficiency-oriented 3DGS~\cite{fan2024lightgaussian, hanson2025pup, liu2025maskgaussian, liu2024compgs}, a natural extension is to explore pruning strategies during training. Yet, when applied to poisoned inputs, these clean-input-based pruning methods become unreliable due to the difficulty of distinguishing fine details from malicious noisy textures, where maintaining Gaussians from attack-induced components is inherently challenging. In addition, since the training supervision is poisoned, images contain noise when guiding the reconstruction, thereby resulting in accuracy degradation. These limitations highlight the need for a principled defense that selectively suppresses poisoned overgrowth without harming the natural structure of the scene. 

In this paper, we explore the underlying perspective beyond spatial appearance. Our insight builds on the observation in~\cref{fig:teaser}(b): poisoning attacks exploit the adaptive growth of 3DGS by injecting stealthy perturbations that, while visually subtle, create distinct spectral discrepancies between poisoned images and clean ones. As observed, poisoned images exhibit abnormal amplification in high-frequency regions and directional anisotropy in the Fourier domain, deviating from the clean distribution. However, these spectral distortions are subtle in pixel space, making spatial detection unreliable. As a result, these spectral distortions drive the optimizer to inflate noisy high-frequency components in an attempt to match the poisoned inputs during training, ultimately causing unstable Gaussian overgrowth. This observation suggests that the root cause of overgrowth lies not in spatial structures, but in spectral behaviors. Motivated by these, we address the attacks from a spectral perspective, identifying abnormally high-frequency responses as underlying driving factors. Yet, directly suppressing high frequencies is insufficient, as natural scene details also contain legitimate high-frequency components, and naive filtering would degrade reconstruction fidelity. Therefore, we examine how stealthy perturbations distort the spectral behavior of the training data, and design a spectral defense under strong poisoning attacks in 3DGS.

Specifically, our spectral defense operates jointly on the 3D Gaussians and 2D renderings in~\cref{fig:teaser}(a). At the 3D level, we introduce a frequency-aware importance measure that links Gaussian covariance to its spectral response, enabling selective filtering of Gaussians exhibiting abnormally high-frequency amplification. This enables 3DGS to adaptively prune redundant splats, thereby suppressing attack-induced overgrowth. At the 2D level, we distinguish and regulate the high-frequency energy by exploiting the observation that clean images exhibit approximately isotropic high-frequency structures with even angular distributions, whereas poisoned renderings introduce anisotropic artifacts. In this way, our regularization penalizes such directional anisotropy and encourages isotropic distributions, allowing the defended spectrum to progressively restore clean spectral behaviors in~\cref{fig:teaser}(b). By preserving naturally isotropic high-frequency structures while suppressing attack-induced anisotropy, our method stabilizes Gaussian growth and enables robust 3DGS reconstruction under poisoning attacks. Our contributions are summarized as follows:

\begin{itemize}

    \item We present the first defense against the resource-targeting attack in 3DGS.

    \item Our spectral defense develops a 3D frequency filter that suppresses Gaussian overgrowth and a 2D spectral regularization that constrains noisy patterns.
    
    \item We conduct experiments on clean, attack, and defense settings, suppressing overgrowth by up to $5.92\times$, reducing peak memory by up to $ 3.66\times$, improving rendering speed by up to $4.34\times$, and preserving rendering quality.
    
\end{itemize}

\section{Related Work}
\label{sec:related_work}
\vspace{-2mm}
\textbf{Efficient 3D Gaussian Splatting.}
3D Gaussian Splatting (3DGS) has rapidly become a main representation for real-time 3D reconstruction and rendering due to its explicit structure, differentiable rasterization, and high rendering fidelity~\cite{kerbl20233d}. The success of 3DGS has spurred widespread adoption across 3D generation~\cite{tang2024lgm, chen2024text}, SLAM and mapping \cite{wen2025segs, homeyer2025droid}, avatar reconstruction \cite{xu2025sequential}, and dynamic scene modeling \cite{li20253d}. Moreover, a series of methods have explored compressed and lightweight 3DGS, where some approaches quantize Gaussian attributes using codebooks~\cite{niedermayr2024compressed, lee2024compact} and spatially neighboring relationships~\cite{chen2024hac, liu2024compgs}. To reduce the model size, a parallel line of work focuses on pruning Gaussian primitives, such as learnable masking strategies~\cite{zhang2024lp, liu2025maskgaussian} and scoring strategies~\cite{fan2024lightgaussian, hanson2025pup, fang2024mini}, to remove redundant Gaussians and achieve more compact representations during training. While these efficiency-oriented strategies are fundamentally designed for compression and acceleration rather than robustness, they lack mechanisms to distinguish adversarially induced Gaussians.

\noindent\textbf{Security Challenges on 3D Representations.}
The security challenges on 3D representations have attracted attention, and studies primarily examine adversarial vulnerabilities in neural radiance fields, showing that subtle perturbations can distort geometry, manipulate semantic attributes, or inject scene-level artifacts~\cite{horvath2023targeted, jiang2024nerfail}. More efforts have explored poisoning-based threats, where carefully crafted inputs can poison the training process or manipulate view-dependent appearance~\cite{huang2024towards, jiang2024ipa}. Within this landscape, 3DGS~\cite{kerbl20233d} inherits new security challenges due to its dense Gaussian primitives in 3D representations. 3DGS is vulnerable to both accuracy-targeting~\cite{zeybey2024gaussian, ke2025stealthattack} and resource-targeting~\cite{lu2024poison} adversaries, motivating the need for secure 3D reconstruction pipelines. Poison-Splat \cite{lu2024poison} presents the first resource-targeting attack on computation cost in 3DGS, showing that poisoned images drastically trigger adversarial over-densification of Gaussians. Existing 3D defense methods do not apply to 3DGS, as they neither address adversarial densification nor operate on Gaussian primitives. Besides, some approaches broadly rely on adversarial training~\cite{kuang2024defense, ruan2023towards}, which requires clean supervision and is incompatible with a practical scenario under poisoned images. The defense research targeting 3D Gaussians remains largely unexplored, and we propose a mechanism for 3DGS that suppresses adversarial attacks.

\section{Preliminary}
\label{sec:method_pre}
\vspace{-2mm}
\noindent\textbf{3D Gaussian Splatting.}
3DGS~\cite{kerbl20233d} uses multi-view images to reconstruct 3D scenes consisting of a set of learnable 3D Gaussians $\mathcal{G}=\{G_i\}$. $\mathcal{G}$ is initially constructed from SfM point clouds~\cite{schonberger2016structure}, and each Gaussian $G_i$ is parameterized by position $\mu_i \in \mathbb{R}^3$, covariance matrix $\Sigma_i \in \mathbb{R}^{3 \times 3}$, color $c_i \in \mathbb{R}^3$, and opacity $\alpha_i \in [0,1]$ to define the Gaussian transparency. To optimize these parameters, 3D Gaussians are rendered as 2D images through $\alpha$-blending~\cite{kerbl20233d}, where the color $\dot{C}$ of each pixel is computed by $N$ overlapping Gaussians in front-to-back depth order, and the product of the opacity represents the influence of previous Gaussians within the same pixel. Based on $\dot{C}$, the rendered image $\dot{V}$ is supervised by the Ground Truth image $V$, where the two sets of images are defined as $\mathcal{\dot{V}}=\{\dot{V}_k\}_{k=1}^{K}$ and $\mathcal{V}=\{V_k\}_{k=1}^{K}$ containing $K$ camera views. 3DGS training minimizes the difference between these two sets, employing $\mathcal{L}_1$ loss and $\mathcal{L}_{\text{D-SSIM}}$ loss (Structural Similarity Index Measure) with a hyperparameter $\lambda$:
\begin{equation}
 \begin{aligned}
  \underset{\mathcal{G}}{\min\text{ }}\mathcal{L}(\mathcal{\dot{V}}, \mathcal{V}) = (1 - \lambda) \mathcal{L}_1(\mathcal{\dot{V}}, \mathcal{V}) + \lambda \mathcal{L}_{\text{D-SSIM}}(\mathcal{\dot{V}}, \mathcal{V}),
 \end{aligned}
 \label{eq:loss-gaussian}
\end{equation}
where adaptive densification is performed to add or remove Gaussians to ensure high-quality reconstruction of 3DGS during training optimization.

\vspace{1mm}
\noindent\textbf{Resource-Targeting Attack of 3DGS.}
A security problem of 3DGS is revealed by recent attack, Poison-splat~\cite{lu2024poison}, which poisons input images for over-consumption, drastically increasing the training time and memory cost, referred to as the resource-targeting attack. The attack disrupts normal 3DGS by addressing a max-min bi-level optimization problem: 
\begin{equation}
 \begin{aligned}
  \mathcal{V}^{p} = \underset{\mathcal{V}^{p}}{\arg \max\text{ }} \mathcal{C}(\mathcal{G}^{\ast}) ~\text{ s.t. } \mathcal{G}^{\ast} = \underset{\mathcal{G}}{\arg \min\text{ }}\mathcal{L}(\mathcal{\dot{V}}, \mathcal{V}^{p}),
 \end{aligned}
 \label{eq:bi-level}
\end{equation}
where $\mathcal{C}(\mathcal{G})$ is the computation cost metric designed by the attack and $\mathcal{G}^{\ast}$ represents the proxy 3DGS model trained on poisoned images $\mathcal{V}^{p} =\{V_k^p\}_{k=1}^{K}$. The attack strength $\epsilon$ has $\epsilon$-ball constraint~\cite{madry2018towards} to balance destructiveness and stealthiness. Through the tailored objective approximation and proxy model rendering from the attack, the victim receives poisoned images and conducts 3DGS training without awareness that the images are poisoned, thus tampering with the training cost. Notably, the attack is effective with full access to 3DGS details, making naive defense methods ineffective against the strong attack. Rather than relying on reactive post-processing defenses, we design a defense-oriented pruning mechanism that proactively regulates Gaussian growth during training.

\section{Spectral Defense for 3DGS}
\label{sec:method_our}
\vspace{-2mm}

Recent attacks on 3DGS exploit its \emph{adaptive complexity}: small and stealthy perturbations increase total variation across views, forcing the optimizer to explain spurious high-frequency artifacts and causing excessive Gaussian growth with little change in visual appearance. Motivated by this failure mode, we design a defense that directly targets the \textbf{frequency behavior} of both the 3D representation and its 2D renderings. At the \textbf{3D level}, we introduce a \emph{frequency-aware importance} that links each Gaussian’s covariance to its high-frequency contribution, enabling selective removal of attack-induced and over-densified splats while retaining genuine geometry. At the \textbf{2D level}, we regularize the angular distribution of high-frequency energy to penalize anisotropic artifacts while preserving natural edges and textures. These components constrain the adversary’s ability to grow poisoned high-frequency structures, stabilizing 3DGS training under strong poisoning.

\vspace{-1mm}
\subsection{3D Frequency Filter}
\label{sec:method_3d_freq}
\vspace{-1mm}

Our first component operates directly in the \emph{3D parameter space} of Gaussians and removes adversarially redundant splats \emph{before} they propagate through the scene representation. Rather than treating pruning as a mere efficiency trick, we use it as a \textbf{security primitive} that suppresses attack-induced structures. Concretely, we link each Gaussian’s covariance to its high-frequency response via a Fourier characterization, define a \emph{frequency-aware importance} for individual splats, and apply a periodic pruning rule during 3DGS training to discard Gaussians with abnormally strong high-frequency responses while largely preserving those that encode genuine geometry.

\vspace{1mm}
\noindent\textbf{Gaussian Frequency Representation.}
For a 3D Gaussian $G$ with covariance matrix $\Sigma$ and center $\mu$, its spatial-domain form $G(\texttt{x})$ has $\Sigma$ encoding both orientation and scale. The spatial variable $\texttt{x} \in \mathbb{R}^3$ and frequency variable $t \in \mathbb{R}^3$ form a pair, and its Fourier response
$\mathcal{F}(t)$ describes how the energy of $G(\texttt{x})$ is distributed over frequencies. The resulting spectrum can be decomposed in terms of amplitude and phase components:
\begin{equation}
  \gamma (t) = (2 \pi)^{\frac{3}{2}} |\Sigma|^{\frac{1}{2}}
     \exp\!\big(-2 \pi^2 t^{\top} \Sigma t\big),~~
  \theta (t) = -2 \pi t^{\top} \mu,
 \label{eq:freq_ap}
\end{equation}
where $\gamma(t)$ is the amplitude attenuation and $\theta(t)$ is the phase shift. Here, $\Sigma$ fully determines that smaller eigenvalues experience weaker attenuation and thus stronger high-frequency response, whereas $\mu$ only induces a linear phase term and does not affect the spectral distribution. Consequently, the frequency characteristics of each Gaussian are entirely governed by its covariance, providing a natural frequency-domain descriptor. Step-by-step equations of $G(\texttt{x})$ and $\mathcal{F}(t)$ with derivations of $\gamma(t)$ and $\theta(t)$ are detailed in the supplementary material.

\vspace{1mm}
\noindent\textbf{Frequency-Aware Scoring for Gaussian Pruning.} 
To quantify the frequency contributions of 3D Gaussians, we construct a scalar scoring mechanism that drives \emph{frequency-aware pruning}, removing redundant high-frequency components while preserving meaningful structural detail. Intuitively, Gaussians with extremely strong high-frequency responses are downweighted, as they often correspond to poisoned regions encouraged by the attack. Using the amplitude term in~(\ref{eq:freq_ap}), we define a high-frequency attenuation score:
\begin{equation}
\mathcal{S}(G)
= \exp(-2\pi^2 t^2 \sigma_{\min}^2),
\label{eq:hf_score}
\end{equation}
where $\sigma_{\min}$ is the smallest eigenvalue of the covariance $\Sigma$, controlling the narrowest spatial spread of the Gaussian: a smaller $\sigma_{\min}$ yields a stronger high-frequency response. We evaluate~(\ref{eq:hf_score}) at a fixed cutoff frequency $t$ to measure the expected high-frequency mass of each Gaussian under a common spectral reference, avoiding ambiguities that can arise from adaptive or scene-dependent thresholds. We then map this attenuation score to a high-frequency importance weight:
\begin{equation}
\mathcal{W}(G)
= \big(1 - \mathcal{S}(G)\big)^{\alpha},
\label{eq:hf_weight}
\end{equation}
where $\alpha > 0$ controls the non-linear amplification. In this formulation, splats with excessively large $\mathcal{S}(G)$, indicating strong high-frequency responses, are assigned low weights, whereas Gaussians contributing moderate and structurally relevant high-frequency contents retain higher weights.

\vspace{-1mm}
\subsection{2D Spectral Regularization}
\label{sec:method_2d_freq}
\vspace{-1mm}

While the \emph{3D frequency filter} suppresses adversarially redundant Gaussians in parameter space, the victim model is still optimized on poisoned views and may converge to renderings with poisoned artifacts. To further stabilize the \emph{image-domain} behavior, we introduce a \emph{2D Spectral Regularization} scheme that operates on the Fourier spectra of rendered images and explicitly controls their \emph{directional high-frequency statistics}. Concretely, this component (i) isolates a \emph{high-frequency band} of interest, (ii) aggregates spectral energy over angular sectors to obtain an \emph{orientation-wise distribution}, and (iii) defines an \emph{entropy-based anisotropy loss} that penalizes over-concentrated directional patterns while leaving natural, isotropic textures largely unaffected. In effect, the regularization acts as a \emph{frequency-domain prior} on the rendered outputs, discouraging attack-induced directional noise and promoting perceptually consistent reconstructions.

\begin{figure}[t]
  \centering
   \includegraphics[width=1.0\linewidth]{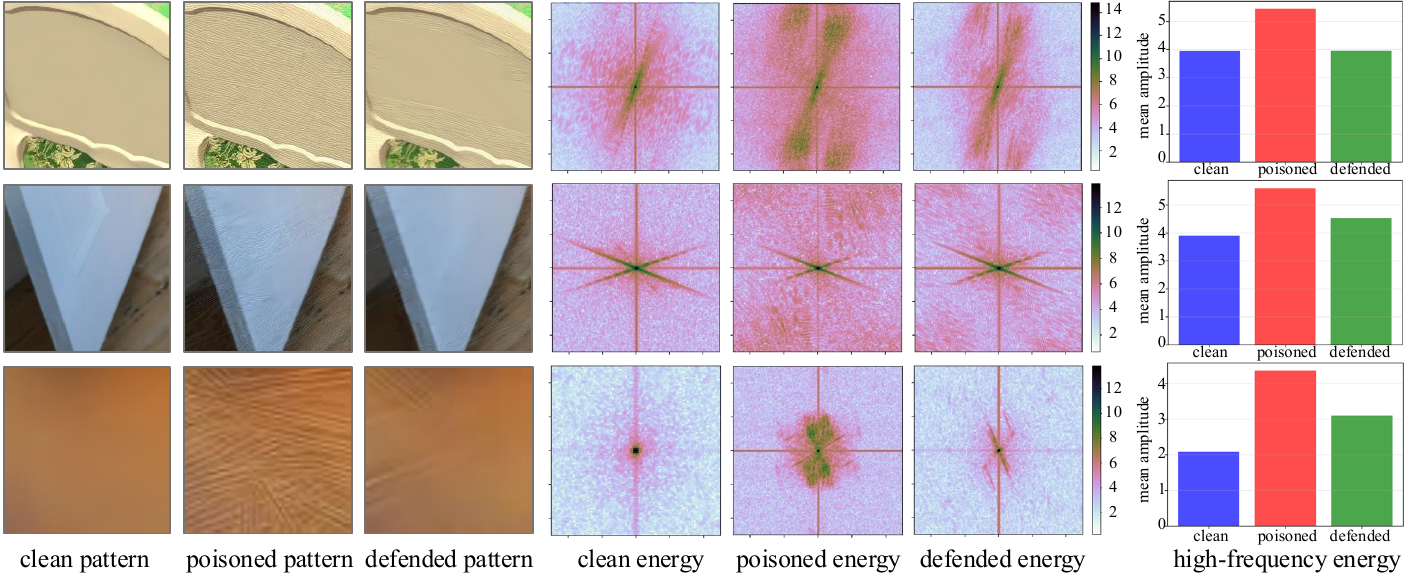}
   \vspace{-6mm}
   \caption{Spectral energy over frequencies under clean, poisoned, and defended settings.}
   \label{fig:prob}
   \vspace{-2mm}
\end{figure}

\vspace{1mm}
\noindent\textbf{Spectral Energy Characterization.}
We analyze the spectral structure of rendered views via a 2D frequency decomposition.  
Given a poisoned image $V^{p}$, its rendered image $\dot V^p$ is produced by differentiable rendering, where each pixel is indexed by spatial coordinates $(h,w)$. We apply a 2D discrete Fourier transform (DFT) to obtain the frequency response $\dot{\mathcal{F}}(u,v)$, where $(u,v)$ denote discrete frequency indices. The complex-valued response $\dot{\mathcal{F}}(u,v)$ can be expressed in terms of amplitude $\dot{\gamma}(u,v)$ and phase $\dot{\theta}(u,v)$. We provide detailed equations in the supplementary material. The amplitude spectrum $\dot{\gamma}(u,v)$ encodes the spectral energy over frequencies in~\cref{fig:prob}. In practice, attack-induced rendering artifacts tend to appear as pronounced responses concentrated along narrow angular directions at higher frequencies, rather than in the low-frequency region. We focus on a band of coefficients corresponding to high-frequency content by thresholds $\dot{\gamma}_{\text{min}}$ and $\dot{\gamma}_{\text{max}}$:
\begin{equation}
\begin{aligned}
\mathcal{E}(u,v)
= \big\{\dot{\mathcal{F}}(u,v)\,\big|\,
\dot \gamma_{\text{min}} \leq \dot \gamma (u,v) \leq \dot \gamma_{\text{max}}\big\}.
\end{aligned}
\label{eq:mask}
\end{equation}
This band-pass selection concentrates on high frequencies with substantial energy and serves as the basis for our subsequent angular energy regularization.

\vspace{1mm}
\noindent\textbf{Angular Distribution.}
To capture the \emph{directional} structure of high-frequency energy $\mathcal{E}(u,v)$, we analyze how this energy is distributed over orientations in the frequency plane, providing a measure of directional smoothness and spectral anisotropy. We discretize the angular domain $[-\pi,\pi)$ into $B$ uniformly spaced bins, and let the $b$-th angular sector be:
\begin{equation}
\begin{aligned}
\tfrac{2\pi b}{B} \;\leq\; \dot{\theta}_b \;<\; \tfrac{2\pi (b+1)}{B},
\end{aligned}
\label{eq:bin}
\end{equation}
where each sector $\dot{\theta}_b$ collects spectral coefficients whose frequency vectors fall within this angular range. For each bin, we aggregate the high-frequency energy from $\mathcal{E}(u,v)$ in~(\ref{eq:mask}) to obtain an \emph{angular energy density} $\mathcal{E}_b$ and a discrete probability distribution $\{\mathcal{P}_b\}_{b=1}^B$ over orientations:
\begin{equation}
\begin{aligned}
\mathcal{E}_b &= \sum_{(u,v) \in \dot{\theta}_b} \mathcal{E}(u,v), \quad
\mathcal{P}_b = \frac{\mathcal{E}_b}{\sum_j \mathcal{E}_j}.
\end{aligned}
\label{eq:bin_norm}
\end{equation}
Intuitively, a clean image in~\cref{fig:teaser}(b) exhibits approximately isotropic content, and the energy spreads relatively evenly across all angular bins, where $\mathcal{P}_b$ is close to flat. In contrast, attack-induced directional artifacts as noisy patterns cause the high frequencies to concentrate within a narrow set of orientations, making $\mathcal{P}_b$ sharply peaked in specific bins with anisotropy content, while our defense restores the distribution similar to the clean
reference.

\begin{algorithm}[t]
\caption{Spectral defense in 3DGS training}
\label{alg:freq_defense}
\begin{algorithmic}[1]
\FOR{iteration $t = 1,2,\dots$}
    \STATE Poisoned views $\mathcal{V}^{p} = \{V_k^{p}\}_{k=1}^{K}$ with camera poses
    \STATE Render $\dot{\mathcal{V}}^{p} = \{\dot V_k^{p}\}_{k=1}^{K}$ from $\mathcal{G}$
    \IF{$t \bmod T_{\text{prune}} = 0$}
        \STATE Randomly sample $K^\ast$ camera views
        \FOR{each $G \in \mathcal{G}$}
            \STATE Compute $\mathcal{W}(G)$ in~(\ref{eq:hf_weight}); Set $\text{score}(G)$ as $\mathcal{W}(G)$ multiplied by $\text{hit}(G)$
        \ENDFOR
        \STATE Prune $\rho$ Gaussians according to $\text{score}(G)$
    \ENDIF
    \STATE Compute anisotropy loss in~(\ref{eq:anisotropy_loss}), reconstruction loss and total variation loss 
    \STATE Update $\mathcal{G}$ by minimizing the total loss in~(\ref{eq:total_loss})
    
\ENDFOR
\end{algorithmic}
\end{algorithm}

\vspace{1mm}
\noindent\textbf{Anisotropy Loss.}
Given the angular distribution, we quantify its directional concentration $\mathcal{H}$ in $[0, \log B]$ and attain its maximum $\log B$ when $\mathcal{P}_b = 1/B$ for all $b$. The $\mathcal{H}$ is maximal when the high-frequency energy is distributed uniformly across all directions defined as isotropic, and minimal when it collapses into sparse dominant orientations defined as anisotropic. To make the regularization comparable across scenes and training iterations, we normalize it by its upper bound $\log B$ to be in $[0,1]$ as:
\begin{equation}
\begin{aligned}
\mathcal{H} = - \sum_{b=1}^{B} \mathcal{P}_b \log \mathcal{P}_b, \quad \mathrm{norm}(\mathcal{H}) = \frac{\mathcal{H}}{\log B},
\end{aligned}
\label{eq:angular_entropy}
\end{equation}
with $1$ for isotropic response and $0$ for anisotropic response with extreme directions. We define the anisotropy loss:
\begin{equation}
\begin{aligned}
\underset{\mathcal{G}}{\min}\mathcal{L}_{\text{ani}}(\dot{\mathcal{V}}^{p})
= 1 - \mathrm{norm}(\mathcal{H}),
\end{aligned}
\label{eq:anisotropy_loss}
\end{equation}
where $\dot{\mathcal{V}}^{p} = \{\dot V_k^{p}\}_{k=1}^{K}$ is the set of rendered poisoned views and the loss is computed per view and averaged over $K$ viewpoints. This regularizer penalizes low-entropy, directionally concentrated spectra, suppressing streaking and banding artifacts without over-smoothing the images.

\vspace{-1mm}
\subsection{Overall Pipeline}
\label{sec:overall}
\vspace{-1mm}

We integrate the 3D frequency filter and image-domain spectral regularizer into a unified training procedure for adversarially robust 3DGS, as in Algorithm~\ref{alg:freq_defense}. 

\vspace{1mm}
\noindent\textbf{Frequency-aware Pruning in Parameter Space.}
Based on the analysis in~\Cref{sec:method_3d_freq}, each Gaussian $G \in \mathcal{G}$ is assigned a high-frequency importance $\mathcal{W}(G)$ via~(\ref{eq:hf_weight}). To incorporate geometric visibility, we project Gaussians into the image plane and count how often they are intersected by training rays, following the ray-hit indicator $\text{hit}(G)$~\cite{fan2024lightgaussian}. Instead of exhaustively using all $K$ views, we randomly sample $K^\ast$ representative camera poses at each pruning step to balance the coverage and computational cost. For each Gaussian, we form a combined $\text{score}(G)$ as $\mathcal{W}(G)\,\cdot\,\text{hit}(G)$, which reflects both its spectral contribution and its spatial support across the scene. Gaussians with low scores are rarely observed yet exhibit strong high-frequency behavior, defined as attack-induced. At a fixed pruning interval $T_{\text{prune}}$ as $100$ by default setting, we sort $\{\text{score}(G)\}_{G\in\mathcal{G}}$ and prune the lowest $\rho$ Gaussians, where $\rho$ is a percent. This periodic pruning removes redundant splats while preserving spectrally important components.

\vspace{1mm}
\noindent\textbf{Spectral Regularization in Image Space.}
In parallel, we enforce the 2D spectral prior from~\Cref{sec:method_2d_freq} at every training iteration. Let $\mathcal{\dot V}^{p} = \{\dot V_k^{p}\}_{k=1}^{K}$ denote the set of rendered poisoned views. For each $\dot V_k^{p}$, we compute the spectrum $\dot{\mathcal{F}}_k$, extract the high-frequency band $\mathcal{E}_k(u,v)$ in~(\ref{eq:mask}), derive the angular distribution $\{\mathcal{P}_{k,b}\}_{b=1}^{B}$ and normalized entropy $\mathrm{norm}(\mathcal{H}_k)$, and then obtain an anisotropy loss $\mathcal{L}_{\text{ani}}(\dot V_k^{p})$ as in~(\ref{eq:anisotropy_loss}). Aggregating over the $K$ rendered views yields the frequency-domain regularizer as:
\begin{equation}
\begin{aligned}
\mathcal{L}_{\text{freq}}(\mathcal{\dot V}^{p})
= \frac{1}{K} \sum_{k=1}^{K} \mathcal{L}_{\text{ani}}(\dot V_k^{p}),
\end{aligned}
\label{eq:loss_freq}
\end{equation}
which penalizes the low-entropy, directionally-biased spectra corresponding to streaking and banding artifacts, while leaving isotropic high-frequency content largely unaffected. We further include a total variation loss $\mathcal{L}_{\text{tv}}(\mathcal{\dot V}^{p})$~\cite{lu2024poison} to promote local smoothness and suppress pixel-level noise.

\vspace{1mm}
\noindent\textbf{Joint Optimization and Training Loop.}
Let $\mathcal{V}^{p}$ denote the target and poisoned views, and $\mathcal{L}(\mathcal{\dot V}^{p}, \mathcal{V}^{p})$ is the standard 3DGS reconstruction loss including the photometric and structural similarity terms. The overall objective combines reconstruction and frequency-domain priors:
\begin{equation}
\underset{\mathcal{G}}{\min}\;
\Big(
\mathcal{L}(\mathcal{\dot V}^{p}, \mathcal{V}^{p})
+ \lambda \big(
\mathcal{L}_{\text{freq}}(\mathcal{\dot V}^{p})
+ \mathcal{L}_{\text{tv}}(\mathcal{\dot V}^{p})
\big)
\Big),
\label{eq:total_loss}
\end{equation}
where $\lambda$ is a loss weight. As in Algorithm~\ref{alg:freq_defense}, each iteration updates the Gaussian set $\mathcal{G}$ by minimizing~(\ref{eq:total_loss}), and every $T_{\text{prune}}$ steps performs frequency-aware pruning based on $\text{score}(G)$. This joint optimization stabilizes the rendered frequency distribution and suppresses anisotropic high-frequency noise, while the 3D frequency filter prevents adversarial densification of splats in parameter space.

\begin{table*}[t]
\tiny
\caption{Training statistics on all scenes from the Tanks and Temples (TT), NeRF-Synthetic (NS), and Mip-NeRF 360 (MIP) datasets under clean, poison, and defense settings. We report the maximum Gaussian number (M), peak GPU memory (MB), and training time (min).}
\vspace{-3mm}
\renewcommand{\arraystretch}{0.9}
\label{tab:exp_num}
  \centering
  \setlength{\tabcolsep}{1.2pt}
  \begin{tabular}{l|ccc|ccc|ccc}
    \toprule
    Metric & \multicolumn{3}{c|}{Max Gaussian Number (M) $\downarrow$}
    & \multicolumn{3}{c|}{Peak GPU Memory (MB) $\downarrow$}
    & \multicolumn{3}{c}{Training Time (min) $\downarrow$} \\
    \midrule 
    Setting & clean & poison & defense
    & clean & poison & defense
    & clean & poison & defense \\
    \midrule

TT-Auditorium & 0.692 & 2.740$_{\textcolor{p}{3.96\!\times\!\uparrow}}$ & 0.907$_{\textcolor{d}{3.02\!\times\!\downarrow}}$ &4918 & 15280$_{\textcolor{p}{3.11\!\times\!\uparrow}}$ & 7633$_{\textcolor{d}{2.00\!\times\!\downarrow}}$ &10.91 & 18.88$_{\textcolor{p}{1.73\!\times\!\uparrow}}$ & 15.27$_{\textcolor{d}{1.24\!\times\!\downarrow}}$ \\

TT-Ballroom & 3.123 & 4.007$_{\textcolor{p}{1.28\!\times\!\uparrow}}$ & 1.983$_{\textcolor{d}{2.02\!\times\!\downarrow}}$ &9902 & 11510$_{\textcolor{p}{1.16\!\times\!\uparrow}}$ & 7075$_{\textcolor{d}{1.63\!\times\!\downarrow}}$ &20.41 & 23.32$_{\textcolor{p}{1.14\!\times\!\uparrow}}$ & 19.81$_{\textcolor{d}{1.18\!\times\!\downarrow}}$ \\

TT-Barn & 0.998 & 1.853$_{\textcolor{p}{1.86\!\times\!\uparrow}}$ & 0.903$_{\textcolor{d}{2.05\!\times\!\downarrow}}$ &6801 & 9594$_{\textcolor{p}{1.41\!\times\!\uparrow}}$ & 6569$_{\textcolor{d}{1.46\!\times\!\downarrow}}$ &11.26 & 14.70$_{\textcolor{p}{1.31\!\times\!\uparrow}}$ & 12.33$_{\textcolor{d}{1.19\!\times\!\downarrow}}$ \\

TT-Caterpillar & 1.272 & 1.877$_{\textcolor{p}{1.48\!\times\!\uparrow}}$ & 0.738$_{\textcolor{d}{2.54\!\times\!\downarrow}}$ &6279 & 7589$_{\textcolor{p}{1.21\!\times\!\uparrow}}$ & 5398$_{\textcolor{d}{1.41\!\times\!\downarrow}}$ &11.83 & 14.02$_{\textcolor{p}{1.19\!\times\!\uparrow}}$ & 13.70$_{\textcolor{d}{1.02\!\times\!\downarrow}}$ \\

TT-Church & 2.312 & 3.342$_{\textcolor{p}{1.45\!\times\!\uparrow}}$ & 1.292$_{\textcolor{d}{2.59\!\times\!\downarrow}}$ &9032 & 11405$_{\textcolor{p}{1.26\!\times\!\uparrow}}$ & 7534$_{\textcolor{d}{1.51\!\times\!\downarrow}}$ &16.86 & 20.24$_{\textcolor{p}{1.20\!\times\!\uparrow}}$ & 16.08$_{\textcolor{d}{1.26\!\times\!\downarrow}}$ \\

TT-Courthouse & 0.600 & 0.726$_{\textcolor{p}{1.21\!\times\!\uparrow}}$ & 0.626$_{\textcolor{d}{1.16\!\times\!\downarrow}}$ &10476 & 12196$_{\textcolor{p}{1.16\!\times\!\uparrow}}$ & 10349$_{\textcolor{d}{1.18\!\times\!\downarrow}}$ &11.33 & 12.02$_{\textcolor{p}{1.06\!\times\!\uparrow}}$ & 11.72$_{\textcolor{d}{1.03\!\times\!\downarrow}}$ \\

TT-Courtroom & 2.950 & 5.403$_{\textcolor{p}{1.83\!\times\!\uparrow}}$ & 1.903$_{\textcolor{d}{2.84\!\times\!\downarrow}}$ &9028 & 14562$_{\textcolor{p}{1.61\!\times\!\uparrow}}$ & 8102$_{\textcolor{d}{1.80\!\times\!\downarrow}}$ &17.70 & 27.67$_{\textcolor{p}{1.56\!\times\!\uparrow}}$ & 22.23$_{\textcolor{d}{1.24\!\times\!\downarrow}}$ \\

TT-Family & 2.119 & 3.696$_{\textcolor{p}{1.74\!\times\!\uparrow}}$ & 1.298$_{\textcolor{d}{2.85\!\times\!\downarrow}}$ &6140 & 9994$_{\textcolor{p}{1.63\!\times\!\uparrow}}$ & 5314$_{\textcolor{d}{1.88\!\times\!\downarrow}}$ &13.49 & 20.27$_{\textcolor{p}{1.50\!\times\!\uparrow}}$ & 16.49$_{\textcolor{d}{1.23\!\times\!\downarrow}}$ \\

TT-Francis & 0.765 & 1.638$_{\textcolor{p}{2.14\!\times\!\uparrow}}$ & 0.705$_{\textcolor{d}{2.32\!\times\!\downarrow}}$ &4835 & 6246$_{\textcolor{p}{1.29\!\times\!\uparrow}}$ & 4713$_{\textcolor{d}{1.33\!\times\!\downarrow}}$ &9.99 & 13.02$_{\textcolor{p}{1.30\!\times\!\uparrow}}$ & 10.65$_{\textcolor{d}{1.22\!\times\!\downarrow}}$ \\

TT-Horse & 1.296 & 2.513$_{\textcolor{p}{1.94\!\times\!\uparrow}}$ & 0.954$_{\textcolor{d}{2.63\!\times\!\downarrow}}$ &4542 & 7159$_{\textcolor{p}{1.58\!\times\!\uparrow}}$ & 4430$_{\textcolor{d}{1.62\!\times\!\downarrow}}$ &11.74 & 16.37$_{\textcolor{p}{1.39\!\times\!\uparrow}}$ & 13.97$_{\textcolor{d}{1.17\!\times\!\downarrow}}$ \\

TT-Ignatius & 3.198 & 3.935$_{\textcolor{p}{1.23\!\times\!\uparrow}}$ & 1.380$_{\textcolor{d}{2.85\!\times\!\downarrow}}$ &9150 & 11115$_{\textcolor{p}{1.21\!\times\!\uparrow}}$ & 6414$_{\textcolor{d}{1.73\!\times\!\downarrow}}$ &17.69 & 20.98$_{\textcolor{p}{1.19\!\times\!\uparrow}}$ & 15.02$_{\textcolor{d}{1.40\!\times\!\downarrow}}$ \\

TT-Lighthouse & 0.834 & 1.144$_{\textcolor{p}{1.37\!\times\!\uparrow}}$ & 0.565$_{\textcolor{d}{2.02\!\times\!\downarrow}}$ &6751 & 8168$_{\textcolor{p}{1.21\!\times\!\uparrow}}$ & 6129$_{\textcolor{d}{1.33\!\times\!\downarrow}}$ &12.03 & 13.17$_{\textcolor{p}{1.09\!\times\!\uparrow}}$ & 12.52$_{\textcolor{d}{1.05\!\times\!\downarrow}}$ \\

TT-M60 & 1.631 & 2.990$_{\textcolor{p}{1.83\!\times\!\uparrow}}$ & 0.946$_{\textcolor{d}{3.16\!\times\!\downarrow}}$ &7375 & 10373$_{\textcolor{p}{1.41\!\times\!\uparrow}}$ & 6787$_{\textcolor{d}{1.53\!\times\!\downarrow}}$ &13.44 & 18.62$_{\textcolor{p}{1.39\!\times\!\uparrow}}$ & 14.32$_{\textcolor{d}{1.30\!\times\!\downarrow}}$ \\

TT-Meetingroom & 1.256 & 2.885$_{\textcolor{p}{2.30\!\times\!\uparrow}}$ & 1.144$_{\textcolor{d}{2.52\!\times\!\downarrow}}$ &6543 & 10108$_{\textcolor{p}{1.54\!\times\!\uparrow}}$ & 6361$_{\textcolor{d}{1.59\!\times\!\downarrow}}$ &12.07 & 18.43$_{\textcolor{p}{1.53\!\times\!\uparrow}}$ & 14.63$_{\textcolor{d}{1.26\!\times\!\downarrow}}$ \\

TT-Museum & 4.470 & 7.220$_{\textcolor{p}{1.62\!\times\!\uparrow}}$ & 2.431$_{\textcolor{d}{2.97\!\times\!\downarrow}}$ &12308 & 18372$_{\textcolor{p}{1.49\!\times\!\uparrow}}$ & 9684$_{\textcolor{d}{1.90\!\times\!\downarrow}}$ &23.13 & 34.06$_{\textcolor{p}{1.47\!\times\!\uparrow}}$ & 28.01$_{\textcolor{d}{1.22\!\times\!\downarrow}}$ \\

TT-Palace & 0.703 & 0.716$_{\textcolor{p}{1.02\!\times\!\uparrow}}$ & 0.570$_{\textcolor{d}{1.26\!\times\!\downarrow}}$ &7154 & 7410$_{\textcolor{p}{1.04\!\times\!\uparrow}}$ & 6447$_{\textcolor{d}{1.15\!\times\!\downarrow}}$ &11.38 & 11.93$_{\textcolor{p}{1.05\!\times\!\uparrow}}$ & 10.21$_{\textcolor{d}{1.17\!\times\!\downarrow}}$ \\

TT-Panther & 1.782 & 3.444$_{\textcolor{p}{1.93\!\times\!\uparrow}}$ & 1.155$_{\textcolor{d}{2.98\!\times\!\downarrow}}$ &7497 & 11290$_{\textcolor{p}{1.51\!\times\!\uparrow}}$ & 7001$_{\textcolor{d}{1.61\!\times\!\downarrow}}$ &13.56 & 20.63$_{\textcolor{p}{1.52\!\times\!\uparrow}}$ & 17.13$_{\textcolor{d}{1.20\!\times\!\downarrow}}$ \\

TT-Playground & 2.279 & 4.262$_{\textcolor{p}{1.87\!\times\!\uparrow}}$ & 1.728$_{\textcolor{d}{2.47\!\times\!\downarrow}}$ &7659 & 11712$_{\textcolor{p}{1.53\!\times\!\uparrow}}$ & 7597$_{\textcolor{d}{1.54\!\times\!\downarrow}}$ &15.85 & 24.23$_{\textcolor{p}{1.53\!\times\!\uparrow}}$ & 19.07$_{\textcolor{d}{1.27\!\times\!\downarrow}}$ \\

TT-Temple & 0.871 & 1.349$_{\textcolor{p}{1.55\!\times\!\uparrow}}$ & 0.652$_{\textcolor{d}{2.07\!\times\!\downarrow}}$ &5840 & 13600$_{\textcolor{p}{2.33\!\times\!\uparrow}}$ & 4887$_{\textcolor{d}{2.78\!\times\!\downarrow}}$ &11.59 & 14.36$_{\textcolor{p}{1.24\!\times\!\uparrow}}$ & 12.32$_{\textcolor{d}{1.17\!\times\!\downarrow}}$ \\

TT-Train & 1.118 & 1.334$_{\textcolor{p}{1.19\!\times\!\uparrow}}$ & 0.500$_{\textcolor{d}{2.67\!\times\!\downarrow}}$ &5674 & 15805$_{\textcolor{p}{2.79\!\times\!\uparrow}}$ & 4324$_{\textcolor{d}{3.66\!\times\!\downarrow}}$ &11.83 & 13.17$_{\textcolor{p}{1.11\!\times\!\uparrow}}$ & 12.68$_{\textcolor{d}{1.04\!\times\!\downarrow}}$ \\

TT-Truck & 2.509 & 3.597$_{\textcolor{p}{1.43\!\times\!\uparrow}}$ & 1.306$_{\textcolor{d}{2.75\!\times\!\downarrow}}$ &7655 & 13304$_{\textcolor{p}{1.74\!\times\!\uparrow}}$ & 6155$_{\textcolor{d}{2.16\!\times\!\downarrow}}$ &15.77 & 21.82$_{\textcolor{p}{1.38\!\times\!\uparrow}}$ & 16.20$_{\textcolor{d}{1.35\!\times\!\downarrow}}$ \\

\textbf{Average} & 1.751 & 2.889$_{\textcolor{p}{1.65\!\times\!\uparrow}}$ & 1.128$_{\textcolor{d}{2.56\!\times\!\downarrow}}$ & 7408 & 11276$_{\textcolor{p}{1.52\!\times\!\uparrow}}$ & 6614$_{\textcolor{d}{1.70\!\times\!\downarrow}}$ & 13.99 & 18.66$_{\textcolor{p}{1.33\!\times\!\uparrow}}$ & 15.45$_{\textcolor{d}{1.21\!\times\!\downarrow}}$
\\

    \midrule

NS-chair & 0.496 & 0.973$_{\textcolor{p}{1.96\!\times\!\uparrow}}$ & 0.281$_{\textcolor{d}{3.46\!\times\!\downarrow}}$ &2848 & 7358$_{\textcolor{p}{2.58\!\times\!\uparrow}}$ & 3963$_{\textcolor{d}{1.86\!\times\!\downarrow}}$ &7.16 & 11.08$_{\textcolor{p}{1.55\!\times\!\uparrow}}$ & 9.23$_{\textcolor{d}{1.20\!\times\!\downarrow}}$ \\

NS-drums & 0.391 & 0.701$_{\textcolor{p}{1.79\!\times\!\uparrow}}$ & 0.234$_{\textcolor{d}{3.00\!\times\!\downarrow}}$ &2538 & 7078$_{\textcolor{p}{2.79\!\times\!\uparrow}}$ & 3061$_{\textcolor{d}{2.31\!\times\!\downarrow}}$ &6.06 & 9.76$_{\textcolor{p}{1.61\!\times\!\uparrow}}$ & 7.71$_{\textcolor{d}{1.27\!\times\!\downarrow}}$ \\

NS-ficus & 0.259 & 0.273$_{\textcolor{p}{1.05\!\times\!\uparrow}}$ & 0.181$_{\textcolor{d}{1.51\!\times\!\downarrow}}$ &2263 & 3120$_{\textcolor{p}{1.38\!\times\!\uparrow}}$ & 2827$_{\textcolor{d}{1.10\!\times\!\downarrow}}$ &5.92 & 6.17$_{\textcolor{p}{1.04\!\times\!\uparrow}}$ & 6.08$_{\textcolor{d}{1.01\!\times\!\downarrow}}$ \\

NS-hotdog & 0.184 & 1.129$_{\textcolor{p}{6.14\!\times\!\uparrow}}$ & 0.304$_{\textcolor{d}{3.71\!\times\!\downarrow}}$ &2318 & 28124$_{\textcolor{p}{12.13\!\times\!\uparrow}}$ & 7781$_{\textcolor{d}{3.61\!\times\!\downarrow}}$ &6.53 & 13.48$_{\textcolor{p}{2.06\!\times\!\uparrow}}$ & 11.03$_{\textcolor{d}{1.22\!\times\!\downarrow}}$ \\

NS-lego & 0.353 & 0.829$_{\textcolor{p}{2.35\!\times\!\uparrow}}$ & 0.287$_{\textcolor{d}{2.89\!\times\!\downarrow}}$ &2668 & 9024$_{\textcolor{p}{3.38\!\times\!\uparrow}}$ & 5179$_{\textcolor{d}{1.74\!\times\!\downarrow}}$ &6.48 & 10.48$_{\textcolor{p}{1.62\!\times\!\uparrow}}$ & 9.19$_{\textcolor{d}{1.14\!\times\!\downarrow}}$ \\

NS-materials & 0.165 & 0.425$_{\textcolor{p}{2.58\!\times\!\uparrow}}$ & 0.219$_{\textcolor{d}{1.94\!\times\!\downarrow}}$ &2214 & 3544$_{\textcolor{p}{1.60\!\times\!\uparrow}}$ & 2798$_{\textcolor{d}{1.27\!\times\!\downarrow}}$ &5.93 & 7.99$_{\textcolor{p}{1.35\!\times\!\uparrow}}$ & 6.82$_{\textcolor{d}{1.17\!\times\!\downarrow}}$ \\

NS-mic & 0.207 & 0.366$_{\textcolor{p}{1.77\!\times\!\uparrow}}$ & 0.299$_{\textcolor{d}{1.22\!\times\!\downarrow}}$ &2361 & 4826$_{\textcolor{p}{2.04\!\times\!\uparrow}}$ & 2958$_{\textcolor{d}{1.63\!\times\!\downarrow}}$ &5.65 & 8.12$_{\textcolor{p}{1.44\!\times\!\uparrow}}$ & 6.90$_{\textcolor{d}{1.18\!\times\!\downarrow}}$ \\

NS-ship & 0.273 & 1.063$_{\textcolor{p}{3.89\!\times\!\uparrow}}$ & 0.380$_{\textcolor{d}{2.80\!\times\!\downarrow}}$ &2892 & 15589$_{\textcolor{p}{5.39\!\times\!\uparrow}}$ & 5167$_{\textcolor{d}{3.02\!\times\!\downarrow}}$ &6.68 & 12.57$_{\textcolor{p}{1.88\!\times\!\uparrow}}$ & 10.87$_{\textcolor{d}{1.16\!\times\!\downarrow}}$ \\

\textbf{Average}
& 0.291 & 0.720$_{\textcolor{p}{2.47\!\times\!\uparrow}}$ & 0.273$_{\textcolor{d}{2.64\!\times\!\downarrow}}$ & 2513 & 9833$_{\textcolor{p}{3.91\!\times\!\uparrow}}$ & 4217$_{\textcolor{d}{2.33\!\times\!\downarrow}}$ & 6.30 & 9.96$_{\textcolor{p}{1.58\!\times\!\uparrow}}$ & 8.48$_{\textcolor{d}{1.17\!\times\!\downarrow}}$
\\

    \midrule

MIP-bicycle & 5.782 & 10.068$_{\textcolor{p}{1.74\!\times\!\uparrow}}$ & 1.874$_{\textcolor{d}{5.37\!\times\!\downarrow}}$ &17533 & 25865$_{\textcolor{p}{1.48\!\times\!\uparrow}}$ & 11201$_{\textcolor{d}{2.31\!\times\!\downarrow}}$ &33.79 & 50.76$_{\textcolor{p}{1.50\!\times\!\uparrow}}$ & 33.18$_{\textcolor{d}{1.53\!\times\!\downarrow}}$ \\

MIP-bonsai & 1.273 & 6.139$_{\textcolor{p}{4.82\!\times\!\uparrow}}$ & 1.037$_{\textcolor{d}{5.92\!\times\!\downarrow}}$ &10067 & 19080$_{\textcolor{p}{1.90\!\times\!\uparrow}}$ & 12182$_{\textcolor{d}{1.57\!\times\!\downarrow}}$ &17.02 & 33.72$_{\textcolor{p}{1.98\!\times\!\uparrow}}$ & 27.85$_{\textcolor{d}{1.21\!\times\!\downarrow}}$ \\

MIP-counter & 1.199 & 4.599$_{\textcolor{p}{3.84\!\times\!\uparrow}}$ & 0.875$_{\textcolor{d}{5.26\!\times\!\downarrow}}$ &10272 & 30332$_{\textcolor{p}{2.95\!\times\!\uparrow}}$ & 11873$_{\textcolor{d}{2.55\!\times\!\downarrow}}$ &19.99 & 34.22$_{\textcolor{p}{1.71\!\times\!\uparrow}}$ & 30.54$_{\textcolor{d}{1.12\!\times\!\downarrow}}$ \\

MIP-flowers & 3.489 & 5.098$_{\textcolor{p}{1.46\!\times\!\uparrow}}$ & 1.335$_{\textcolor{d}{3.82\!\times\!\downarrow}}$ &11356 & 15073$_{\textcolor{p}{1.33\!\times\!\uparrow}}$ & 8150$_{\textcolor{d}{1.85\!\times\!\downarrow}}$ &24.07 & 30.54$_{\textcolor{p}{1.27\!\times\!\uparrow}}$ & 28.46$_{\textcolor{d}{1.07\!\times\!\downarrow}}$ \\

MIP-garden & 5.698 & 7.724$_{\textcolor{p}{1.36\!\times\!\uparrow}}$ & 1.691$_{\textcolor{d}{4.57\!\times\!\downarrow}}$ &16896 & 22238$_{\textcolor{p}{1.32\!\times\!\uparrow}}$ & 9668$_{\textcolor{d}{2.30\!\times\!\downarrow}}$ &34.13 & 45.49$_{\textcolor{p}{1.33\!\times\!\uparrow}}$ & 28.54$_{\textcolor{d}{1.59\!\times\!\downarrow}}$ \\

MIP-kitchen & 1.755 & 6.442$_{\textcolor{p}{3.67\!\times\!\uparrow}}$ & 1.714$_{\textcolor{d}{3.76\!\times\!\downarrow}}$ &11088 & 20864$_{\textcolor{p}{1.88\!\times\!\uparrow}}$ & 11840$_{\textcolor{d}{1.76\!\times\!\downarrow}}$ &22.07 & 42.51$_{\textcolor{p}{1.93\!\times\!\uparrow}}$ & 31.78$_{\textcolor{d}{1.34\!\times\!\downarrow}}$ \\

MIP-room & 1.536 & 7.269$_{\textcolor{p}{4.73\!\times\!\uparrow}}$ & 1.966$_{\textcolor{d}{3.70\!\times\!\downarrow}}$ &11502 & 45749$_{\textcolor{p}{3.98\!\times\!\uparrow}}$ & 14626$_{\textcolor{d}{3.13\!\times\!\downarrow}}$ &20.43 & 48.01$_{\textcolor{p}{2.35\!\times\!\uparrow}}$ & 35.77$_{\textcolor{d}{1.34\!\times\!\downarrow}}$ \\

MIP-stump & 4.491 & 10.232$_{\textcolor{p}{2.28\!\times\!\uparrow}}$ & 3.503$_{\textcolor{d}{2.92\!\times\!\downarrow}}$ &12937 & 24784$_{\textcolor{p}{1.92\!\times\!\uparrow}}$ & 12367$_{\textcolor{d}{2.00\!\times\!\downarrow}}$ &26.90 & 46.36$_{\textcolor{p}{1.72\!\times\!\uparrow}}$ & 37.34$_{\textcolor{d}{1.24\!\times\!\downarrow}}$ \\

MIP-treehill & 3.495 & 5.838$_{\textcolor{p}{1.67\!\times\!\uparrow}}$ & 2.889$_{\textcolor{d}{2.02\!\times\!\downarrow}}$ &10943 & 16022$_{\textcolor{p}{1.46\!\times\!\uparrow}}$ & 11514$_{\textcolor{d}{1.39\!\times\!\downarrow}}$ &23.71 & 33.47$_{\textcolor{p}{1.41\!\times\!\uparrow}}$ & 28.73$_{\textcolor{d}{1.16\!\times\!\downarrow}}$ \\

\textbf{Average} & 3.191 & 7.045$_{\textcolor{p}{2.21\!\times\!\uparrow}}$ & 1.876$_{\textcolor{d}{3.76\!\times\!\downarrow}}$ & 12510 & 24445$_{\textcolor{p}{1.95\!\times\!\uparrow}}$ & 11491$_{\textcolor{d}{2.13\!\times\!\downarrow}}$ & 24.68 & 40.56$_{\textcolor{p}{1.64\!\times\!\uparrow}}$ & 31.35$_{\textcolor{d}{1.29\!\times\!\downarrow}}$ \\

    \bottomrule
  \end{tabular}
  \vspace{-4mm}
\end{table*}

\section{Experiments}
\label{sec:exp}
\vspace{-2mm}


\textbf{Datasets and Evaluation Metrics.}
We conduct experiments on three 3D datasets: Tanks and Temples~\cite{knapitsch2017tanks}, NeRF-Synthetic~\cite{mildenhall2021nerf}, and Mip-NeRF360~\cite{barron2022mip}. Tanks and Temples (TT) dataset has $21$ realistic 3D captures of both indoor and outdoor environments. NeRF-Synthetic (NS) consists of $8$ synthetic objects rendered from multiple viewpoints for synthetic 3D reconstruction. Mip-NeRF360 (MIP) includes $9$ real-world scenes with complex central regions and detailed surrounding backgrounds, providing a more challenging setup for large-scale reconstruction. All datasets are sampled with seven evenly distributed views for training and others for testing from every eight-view clip. We report computation efficiency and reconstruction accuracy. Compared with clean data training, the poisoned data training has larger increases in the number of Gaussians, higher GPU memory consumption, and extended training time. Thus, our defense approach primarily reduces the computational cost of 3DGS in the context of computation cost attacks. We report the rendering speed by Frames Per Second (FPS), Peak Signal-to-Noise Ratio (PSNR), and Structural Similarity Index Measure (SSIM) for rendering quality.

\begin{figure*}[t]
  \centering
   \includegraphics[width=1.0\linewidth]{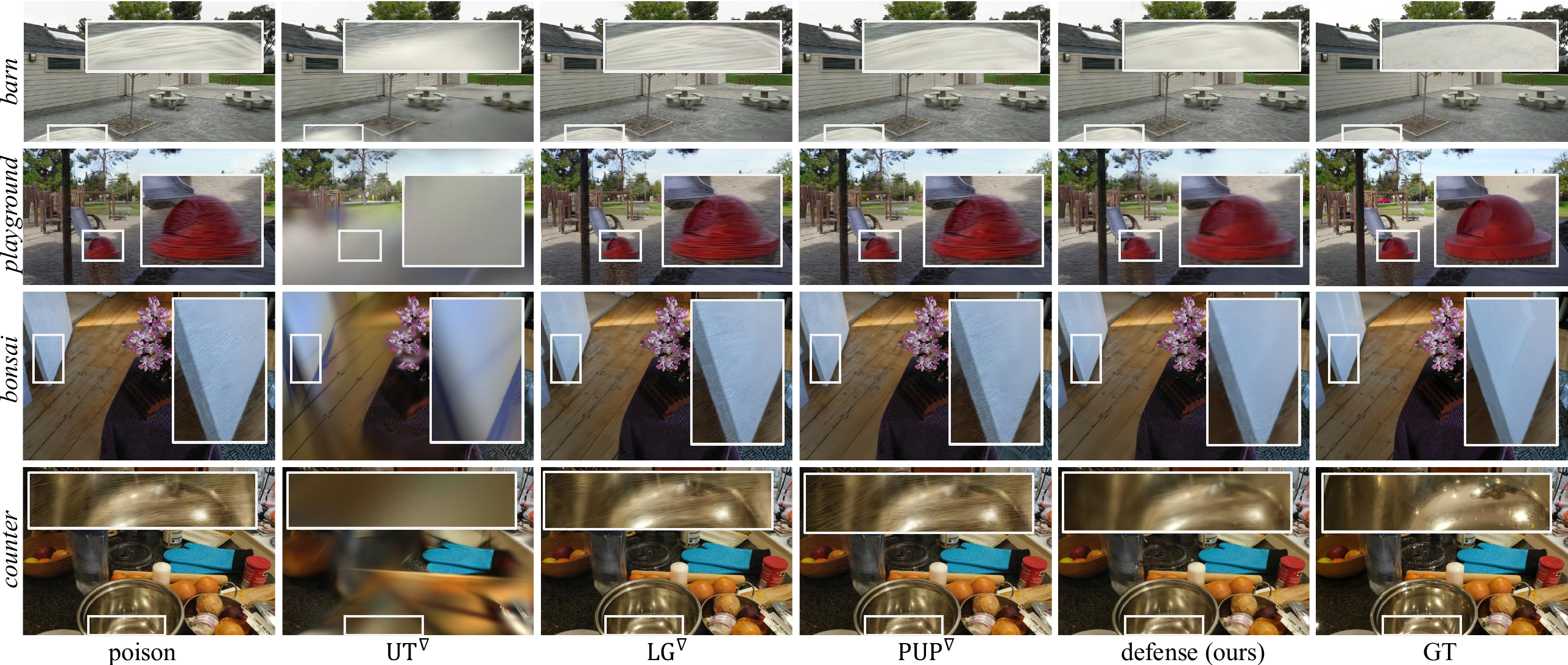}
   \vspace{-6mm}
   \caption{Qualitative comparison under the poison setting across representative scenes.}
   \label{fig:vis}
   \vspace{-2mm}
\end{figure*}

\begin{table*}[t]
\scriptsize
  \caption{Rendering performance on partial scenes from the TT, NS, and MIP datasets, where the results of all scenes are provided in the supplementary materials. We use SSIM and PSNR to measure the visual fidelity of reconstruction compared with various pruning methods, which are retrained based on poisoned images and evaluated on clean images, including $\text{UT}^{\triangledown}$, $\text{LG}^{\triangledown}$, $\text{PUP}^{\triangledown}$, and defense (ours).}
  \vspace{-3mm}
  \label{tab:exp_ssim_psnr}
  \renewcommand{\arraystretch}{0.9}
  \centering
  \setlength{\tabcolsep}{1.0pt}
  \begin{tabular}{l|ccccc|ccccc}
    \toprule
    Metric
    & \multicolumn{5}{c|}{SSIM $\uparrow$}
    & \multicolumn{5}{c}{PSNR $\uparrow$} \\
    \midrule
    Setting & poison & $\text{UT}^{\triangledown}$~\cite{lu2024poison} & $\text{LG}^{\triangledown}$~\cite{fan2024lightgaussian} & $\text{PUP}^{\triangledown}$~\cite{hanson2025pup} & defense
    & poison & $\text{UT}^{\triangledown}$~\cite{lu2024poison} & $\text{LG}^{\triangledown}$~\cite{fan2024lightgaussian} & $\text{PUP}^{\triangledown}$~\cite{hanson2025pup} & defense \\
    \midrule

TT-Barn & 0.64 & 0.68 & 0.65 & 0.64 & \textbf{0.71} & 25.29 & 24.08 & 25.03 & 25.14 & \textbf{25.90} \\

TT-Church & 0.65 & 0.64 & 0.68 & 0.64 & \textbf{0.70} &22.52 & 18.77 & 22.45 & 21.56 & \textbf{22.82} \\

TT-Panther & 0.60 & 0.60 & 0.61 & 0.61 & \textbf{0.69} &25.25 & 17.02 & 25.09 & 24.57 & \textbf{25.79} \\

\midrule

NS-chair & 0.22 & 0.22  & 0.22 & \textbf{0.23} & \textbf{0.23} & 24.60  & 24.57 & 24.93 & 24.82 & \textbf{25.00} \\

NS-hotdog & 0.26 & 0.26 &  0.27 & 0.26 & \textbf{0.28} & 27.00 & 27.08 &  27.02 & 26.82 & \textbf{27.66} \\

NS-materials & 0.22 & 0.21 &  0.22 & 0.22 & \textbf{0.23} & 25.93 & 25.86 &  25.92 & 25.91 & \textbf{26.02} \\

\midrule

MIP-bonsai & 0.64 & 0.75 & 0.67 & 0.69 & \textbf{0.84} & 27.14 & 22.68 & 27.05 & 26.81 & \textbf{29.07}\\

MIP-garden & 0.65 & 0.47 & 0.64 & 0.64 & \textbf{0.73} & 24.63 & 18.99 & 24.05 & 23.77 & \textbf{25.13} \\

MIP-treehill & 0.54 & 0.44 & 0.54 & 0.54 & \textbf{0.57} & 22.32 & 18.26 & 22.34 & 22.15 & \textbf{22.65} \\

    \bottomrule
  \end{tabular}
  \vspace{-4mm}
\end{table*}

\begin{table}[t]
\tiny
\centering
\begin{minipage}{0.49\textwidth}
\centering
\caption{Rendering speed of baselines and our defense method under poison setting.}
\vspace{-3mm}
\label{tab:exp_fps}
\renewcommand{\arraystretch}{0.9}
\setlength{\tabcolsep}{2.2pt}
\begin{tabular}{l|cccc}
\toprule
Metric 
& \multicolumn{4}{c}{FPS $\uparrow$}  \\
\midrule
Setting & poison & $\text{LG}^{\triangledown}$~\cite{fan2024lightgaussian} & $\text{PUP}^{\triangledown}$~\cite{hanson2025pup} & defense \\
\midrule
TT-Barn        & 120 & 199$_{\textcolor{d}{1.66\!\times\!\uparrow}}$ & 184$_{\textcolor{d}{1.53\!\times\!\uparrow}}$ & 236$_{\textcolor{d}{1.97\!\times\!\uparrow}}$ \\
TT-Church      & 94 & 138$_{\textcolor{d}{1.47\!\times\!\uparrow}}$ & 144$_{\textcolor{d}{1.53\!\times\!\uparrow}}$ & 220$_{\textcolor{d}{2.36\!\times\!\uparrow}}$ \\
TT-Panther & 88 & 178$_{\textcolor{d}{2.03\!\times\!\uparrow}}$ & 157$_{\textcolor{d}{1.80\!\times\!\uparrow}}$ & 194$_{\textcolor{d}{2.22\!\times\!\uparrow}}$  \\
NS-chair & 124 & 416$_{\textcolor{d}{3.36\!\times\!\uparrow}}$ & 400$_{\textcolor{d}{3.23\!\times\!\uparrow}}$ & 455$_{\textcolor{d}{3.67\!\times\!\uparrow}}$ \\
NS-hotdog & 91 & 196$_{\textcolor{d}{2.16\!\times\!\uparrow}}$ & 235$_{\textcolor{d}{2.59\!\times\!\uparrow}}$ & 252$_{\textcolor{d}{2.77\!\times\!\uparrow}}$  \\
NS-materials & 132 & 341$_{\textcolor{d}{2.59\!\times\!\uparrow}}$ & 275$_{\textcolor{d}{2.09\!\times\!\uparrow}}$ & 357$_{\textcolor{d}{2.71\!\times\!\uparrow}}$  \\
MIP-bicycle & 38 & 88$_{\textcolor{d}{2.34\!\times\!\uparrow}}$ & 104$_{\textcolor{d}{2.74\!\times\!\uparrow}}$ & 107$_{\textcolor{d}{2.84\!\times\!\uparrow}}$  \\
MIP-garden & 48 & 178$_{\textcolor{d}{3.73\!\times\!\uparrow}}$ & 187$_{\textcolor{d}{3.91\!\times\!\uparrow}}$ & 208$_{\textcolor{d}{4.34\!\times\!\uparrow}}$  \\
MIP-treehill & 58 & 158$_{\textcolor{d}{2.75\!\times\!\uparrow}}$ & 168$_{\textcolor{d}{2.91\!\times\!\uparrow}}$ & 206$_{\textcolor{d}{3.58\!\times\!\uparrow}}$  \\
\bottomrule
\end{tabular}
\end{minipage}
\hfill
\begin{minipage}{0.48\textwidth}
\centering
\caption{The effects on clean inputs. We use clean (def.) as clean with defense.}
\vspace{-3mm}
\label{tab:exp_clean}
\renewcommand{\arraystretch}{0.9}
\setlength{\tabcolsep}{2.3pt}
\begin{tabular}{l|cc|cc}
\toprule
Metric 
& \multicolumn{2}{c|}{NUM (M) $\downarrow$} & \multicolumn{2}{c}{GPU (MB) $\downarrow$}  \\
\midrule
Setting & clean & clean (def.) & clean & clean (def.) \\
\midrule
MIP-bicycle & 5.782 & 1.339$_{\textcolor{d}{4.32\!\times\!\downarrow}}$ &17533 & 9583$_{\textcolor{d}{1.83\!\times\!\downarrow}}$ \\
MIP-bonsai & 1.273 & 0.594$_{\textcolor{d}{2.14\!\times\!\downarrow}}$ &10067 & 9594$_{\textcolor{d}{1.05\!\times\!\downarrow}}$ \\
MIP-counter & 1.199 & 0.552$_{\textcolor{d}{2.17\!\times\!\downarrow}}$ &10272 & 8534$_{\textcolor{d}{1.20\!\times\!\downarrow}}$ \\
MIP-flowers & 3.489 & 1.086$_{\textcolor{d}{3.21\!\times\!\downarrow}}$ &11356 & 7543$_{\textcolor{d}{1.51\!\times\!\downarrow}}$ \\
MIP-garden & 5.698 & 1.386$_{\textcolor{d}{4.11\!\times\!\downarrow}}$ &16896 & 8478$_{\textcolor{d}{1.99\!\times\!\downarrow}}$ \\
MIP-kitchen & 1.755 & 0.613$_{\textcolor{d}{2.86\!\times\!\downarrow}}$ &11088 & 10083$_{\textcolor{d}{1.10\!\times\!\downarrow}}$ \\
MIP-room & 1.536 & 0.840$_{\textcolor{d}{1.83\!\times\!\downarrow}}$ &11502 & 10829$_{\textcolor{d}{1.06\!\times\!\downarrow}}$ \\
MIP-stump & 4.491 & 1.126$_{\textcolor{d}{3.99\!\times\!\downarrow}}$ &12937 & 6733$_{\textcolor{d}{1.92\!\times\!\downarrow}}$ \\
MIP-treehill & 3.495 & 1.018$_{\textcolor{d}{3.43\!\times\!\downarrow}}$ &10943 & 6754$_{\textcolor{d}{1.62\!\times\!\downarrow}}$ \\
\bottomrule
\end{tabular}
\end{minipage}
\vspace{-3mm}
\end{table}

\begin{table*}[t]
\tiny
\caption{Black-box attack results on Scaffold-GS~\cite{lu2024scaffold} as the victim system to demonstrate our defense generalization ability against unknown black-box victim systems.}
\vspace{-3mm}
\renewcommand{\arraystretch}{0.9}
\label{tab:exp_blackbox}
  \centering
  \setlength{\tabcolsep}{2.0pt}
  \begin{tabular}{l|ccc|ccc|ccc}
    \toprule
    Metric & \multicolumn{3}{c|}{Max Gaussian Number (M) $\downarrow$}
    & \multicolumn{3}{c|}{Peak GPU Memory (MB) $\downarrow$}
    & \multicolumn{3}{c}{Training Time (min) $\downarrow$} \\
    \midrule 
    Setting & clean & poison & defense
    & clean & poison & defense
    & clean & poison & defense \\
    \midrule

MIP-bicycle & 9.205 & 17.401$_{\textcolor{p}{1.89\!\times\!\uparrow}}$ & 4.223$_{\textcolor{d}{4.12\!\times\!\downarrow}}$ &15058 & 18261$_{\textcolor{p}{1.21\!\times\!\uparrow}}$ & 12101$_{\textcolor{d}{1.51\!\times\!\downarrow}}$ &35.02 & 40.05$_{\textcolor{p}{1.14\!\times\!\uparrow}}$ & 34.42$_{\textcolor{d}{1.16\!\times\!\downarrow}}$ \\

MIP-bonsai & 4.231 & 11.108$_{\textcolor{p}{2.63\!\times\!\uparrow}}$ & 2.011$_{\textcolor{d}{5.52\!\times\!\downarrow}}$ &9979 & 13959$_{\textcolor{p}{1.40\!\times\!\uparrow}}$ & 8294$_{\textcolor{d}{1.68\!\times\!\downarrow}}$ &29.65 & 34.60$_{\textcolor{p}{1.17\!\times\!\uparrow}}$ & 29.41$_{\textcolor{d}{1.18\!\times\!\downarrow}}$ \\

MIP-counter & 2.855 & 6.506$_{\textcolor{p}{2.28\!\times\!\uparrow}}$ & 1.738$_{\textcolor{d}{3.74\!\times\!\downarrow}}$ &11549 & 20755$_{\textcolor{p}{1.80\!\times\!\uparrow}}$ & 9575$_{\textcolor{d}{2.17\!\times\!\downarrow}}$ &34.56 & 38.63$_{\textcolor{p}{1.12\!\times\!\uparrow}}$ & 32.76$_{\textcolor{d}{1.18\!\times\!\downarrow}}$ \\

MIP-flowers & 7.075 & 11.361$_{\textcolor{p}{1.61\!\times\!\uparrow}}$ & 5.197$_{\textcolor{d}{2.19\!\times\!\downarrow}}$ &9611 & 12603$_{\textcolor{p}{1.31\!\times\!\uparrow}}$ & 6230$_{\textcolor{d}{2.02\!\times\!\downarrow}}$ &31.96 & 34.66$_{\textcolor{p}{1.08\!\times\!\uparrow}}$ & 32.03$_{\textcolor{d}{1.08\!\times\!\downarrow}}$ \\

MIP-garden & 7.587 & 11.795$_{\textcolor{p}{1.55\!\times\!\uparrow}}$ & 4.201$_{\textcolor{d}{2.81\!\times\!\downarrow}}$ &10449 & 15775$_{\textcolor{p}{1.51\!\times\!\uparrow}}$ & 8437$_{\textcolor{d}{1.87\!\times\!\downarrow}}$ &33.46 & 40.15$_{\textcolor{p}{1.20\!\times\!\uparrow}}$ & 33.89$_{\textcolor{d}{1.18\!\times\!\downarrow}}$ \\

MIP-kitchen & 3.303 & 6.632$_{\textcolor{p}{2.01\!\times\!\uparrow}}$ & 2.218$_{\textcolor{d}{2.99\!\times\!\downarrow}}$ &12588 & 16900$_{\textcolor{p}{1.34\!\times\!\uparrow}}$ & 11784$_{\textcolor{d}{1.43\!\times\!\downarrow}}$ &35.70 & 40.88$_{\textcolor{p}{1.15\!\times\!\uparrow}}$ & 33.56$_{\textcolor{d}{1.22\!\times\!\downarrow}}$ \\

MIP-room & 2.747 & 10.247$_{\textcolor{p}{3.73\!\times\!\uparrow}}$ & 3.332$_{\textcolor{d}{3.08\!\times\!\downarrow}}$ &13043 & 14612$_{\textcolor{p}{1.12\!\times\!\uparrow}}$ & 13162$_{\textcolor{d}{1.11\!\times\!\downarrow}}$ &30.97 & 40.42$_{\textcolor{p}{1.31\!\times\!\uparrow}}$ & 33.21$_{\textcolor{d}{1.22\!\times\!\downarrow}}$ \\

MIP-stump & 7.013 & 15.681$_{\textcolor{p}{2.24\!\times\!\uparrow}}$ & 5.400$_{\textcolor{d}{2.90\!\times\!\downarrow}}$ &9047 & 15539$_{\textcolor{p}{1.72\!\times\!\uparrow}}$ & 7838$_{\textcolor{d}{1.98\!\times\!\downarrow}}$ &29.71 & 34.86$_{\textcolor{p}{1.17\!\times\!\uparrow}}$ & 32.85$_{\textcolor{d}{1.06\!\times\!\downarrow}}$ \\

MIP-treehill & 7.821 & 12.260$_{\textcolor{p}{1.57\!\times\!\uparrow}}$ & 6.142$_{\textcolor{d}{2.00\!\times\!\downarrow}}$ &9639 & 12809$_{\textcolor{p}{1.33\!\times\!\uparrow}}$ & 7886$_{\textcolor{d}{1.62\!\times\!\downarrow}}$ &31.08 & 36.53$_{\textcolor{p}{1.18\!\times\!\uparrow}}$ & 31.93$_{\textcolor{d}{1.14\!\times\!\downarrow}}$ \\

    \bottomrule
  \end{tabular}
  \vspace{-4mm}
\end{table*}

\vspace{1mm}
\noindent\textbf{Implementation Details.}
Our defense experiments are conducted based on the official implementation of Poison-splat~\cite{lu2024poison}, which simulates victim behavior from 3D Gaussian Splatting~\cite{kerbl20233d} with designed attacks. We compare a straightforward defense method of imposing universal Gaussian thresholds~\cite{lu2024poison}, and conduct 3DGS pruning methods~\cite{hanson2025pup, fan2024lightgaussian} on the poisoned data training. We adopt the default hyper-parameters provided in the original implementation, as they are demonstrated to be effective across diverse datasets and scenes. The attack strength $\epsilon$ is constrained to $16/255$ by default. We set $t$ as $8$ and $\alpha$ as $2$. We set the view sampling number $K^\ast$ as $48$ and with bin number $B$ as $36$ for 2D frequency regularization. The frequency thresholds are $\dot \gamma_{\text{min}}$ as $0.3$ and $\dot \gamma_{\text{max}}$ as $0.9$. In addition, small scenes from the NS dataset benefit from a mild pruning ratio of $\rho$ as $3.0\%$ and loss weight $\lambda$ as $4$. Large scenes from TT dataset has $\rho$ as $4.5\%$ with $\lambda$ as $4$, and complex scenes from MIP dataset has $\rho$ as $5.0\%$ with $\lambda$ as $5$. All experiments are conducted on a single NVIDIA RTX A6000 GPU.

\vspace{-2mm}
\subsection{Main Results}
\vspace{-2mm}
We evaluate the training cost, accuracy, and rendering efficiency for novel view synthesis against the attack. Experiments are based on clean, poison, and defense settings across indoor and outdoor scenes on three datasets, including Tanks and Temples (TT), NeRF-Synthetic (NS), and Mip-NeRF360 (MIP). We compare our defense with other efficiency-oriented baselines, including universal threshold to control Gaussian number~\cite{lu2024poison} as $\text{UT}^{\triangledown}$, LightGaussian~\cite{fan2024lightgaussian} as $\text{LG}^{\triangledown}$, and PUP 3D-GS~\cite{hanson2025pup} as $\text{PUP}^{\triangledown}$, where $^{\triangledown}$ denotes that we implement these methods under the poison setting. All results are conducted by averaging three individual runs.

\vspace{1mm}
\noindent\textbf{Quantitative Evaluation.}
The quantitative results on the indoor and outdoor scenes are summarized in~\Cref{tab:exp_num}, \Cref{tab:exp_ssim_psnr}, and~\Cref{tab:exp_fps}. By evaluating training statistics, the main performance is assessed on the computation cost in \Cref{tab:exp_num}, reporting Gaussian number, GPU memory usage, and training time. Across all datasets, the poison setting exhibits a substantial increase in the Gaussian number and GPU memory, leading to unstable convergence and over-reconstructed densification in 3DGS. Our method effectively suppresses this redundancy, achieving obvious reductions to nearly restore the clean setting to indicate our defense efficiency. In addition, the~\Cref{tab:exp_ssim_psnr} further details the accuracy under different approaches, where we evaluate the rendering performance in terms of SSIM and PSNR during the test. Specifically, the naive Gaussian number defense $\text{UT}^{\triangledown}$ has a universal threshold of $0.8$M by default. We also implement other pruning baselines containing $\text{LG}^{\triangledown}$ and $\text{PUP}^{\triangledown}$ by following their default settings. Because the victim model is trained on the poison setting and has no access to clean inputs, our defense method aims to keep crucial Gaussian contents through the 3D frequency filter. Our approach effectively suppresses the noise textures introduced during poisoning, enabling the model to reconstruct clean details even without clean supervision. This is reflected by consistently higher PSNR, such as $25.00$ on NS-chair and $29.07$ on MIP-bonsai. Meanwhile, the restored Gaussian sparsity also enhances rendering speed, and we report corresponding FPS in~\Cref{tab:exp_fps}. Our method consistently increases rendering speed over the poisoned setting, achieving FPS improvements ranging from moderate gains to substantial boosts, such as TT-Barn from $120$ to $236$. We achieve higher improvements compared with two pruning baselines, which also partially recover the rendering performance. In addition, we evaluate our defense effects under clean inputs and black-box attacks in~\Cref{tab:exp_clean} and~\Cref{tab:exp_blackbox}, respectively. \Cref{tab:exp_clean} reports defense effects under clean inputs to demonstrate our competitive rendering quality. \Cref{tab:exp_blackbox} reports black-box attack results on Scaffold-GS~\cite{lu2024scaffold} as the victim system, which can further demonstrate the robust generalization ability of our defense method under poison setting against unknown black-box victim systems. Overall, our method filters adversarially redundant Gaussians, confirming its effectiveness in reducing redundant Gaussians and restoring efficient rendering. The complete results revolving SSIM, PSNR, and FPS across all scenes on poison inputs, and the defense effects under clean inputs and black-box attacks are provided in our supplementary materials.

\vspace{1mm}
\noindent\textbf{Qualitative Visualization.}
In~\cref{fig:vis}, we observe that existing pruning approaches retain strong high-frequency noise to lose fine details based on TT-Barn, TT-Playground, MIP-bonsai, and MIP-counter datasets. In contrast, our spectral defense effectively removes poisoned patterns and restores natural edges, yielding reconstructions visually closer to the ground truth (GT) across scenes.


\vspace{-2mm}
\subsection{Ablations}
\vspace{-1mm}
We conduct ablations to investigate the effects of design choices based on NS-chair, TT-Barn, and MIP-bonsai datasets. We analyze the evaluation metrics, including NUM as the Gaussian number, GPU as the peak computational memory, FPS as rendering speed, and SSIM and PSNR as visual quality.

\begin{table}[t]
\scriptsize
\centering
\begin{minipage}{0.44\textwidth}
\centering
\caption{Ablation on the reference frequency $t$ with the exponent $\alpha$.}
\vspace{-3mm}
\renewcommand{\arraystretch}{0.9}
\label{tab:ablation_3d_filter}
\setlength{\tabcolsep}{1.2pt}
\begin{tabular}{ll|ccccc}
\toprule
\multicolumn{2}{l|}{NS-chair} & NUM & GPU & FPS & SSIM & PSNR \\
\midrule
$t=4$  & $\alpha=4$ & 0.295 & 3985 & 460 & 0.21 & 24.85  \\
$t=6$  & $\alpha=3$ & 0.306 & 4015 & 434 & 0.21 & 24.92 \\
\rowcolor{black!5} $t=8$  & $\alpha=2$ & 0.281 & 3963 & 455 &  0.23 & 25.00  \\ 
$t=10$  & $\alpha=1$ & 0.272 & 3882 & 449 &  0.22  & 24.43 \\
\bottomrule
\end{tabular}
\end{minipage}
\hfill  
\begin{minipage}{0.53\textwidth}
\centering
\caption{Ablation on the prune ratio $\rho$ with the view sampling number $K^\ast$.}
\vspace{-3mm}
\renewcommand{\arraystretch}{0.9}
\label{tab:ablation_3d_prune}
\setlength{\tabcolsep}{1.3pt}
\begin{tabular}{l|ccccc}
\toprule
NS-chair & NUM & GPU & FPS & SSIM & PSNR \\
\midrule
Pruning 2.0$\%$ with 32 & 0.339 & 4298 &  423 & 0.23 & 24.87 \\
\rowcolor{black!5} Pruning 3.0$\%$ with 48 & 0.281 & 3963 & 455 &  0.23 & 25.00  \\ 
Pruning 4.0$\%$ with 56 & 0.209 & 3721 & 450 & 0.22 & 24.91 \\
Pruning 5.0$\%$ with 72 &  0.155 & 3698 & 467 & 0.21 & 24.54 \\
\bottomrule
\end{tabular}
\end{minipage}
\vspace{-3mm}
\end{table}

\begin{table}[t]
\scriptsize
\centering
\begin{minipage}{0.53\textwidth}
\centering
\caption{Ablation on the thresholds through lower bound $\dot \gamma_{\text{min}}$ and upper bound $\dot \gamma_{\text{max}}$.}
\vspace{-3mm}
\renewcommand{\arraystretch}{0.9}
\label{tab:ablation_2d_gamma}
\setlength{\tabcolsep}{0.4pt}
\begin{tabular}{l|c>{\columncolor{black!5}}ccc}
\toprule
TT-Barn & $[0.28, 0.92]$ & $[0.30, 0.90]$ &  $[0.35, 0.85]$ & $[0.40, 0.85]$ \\
\midrule
NUM & 0.922 & 0.903 & 0.908 &  0.953  \\
GPU & 6834 & 6569 & 6613 & 7002 \\
PSNR & 25.75 & 25.90 &  25.73 & 25.88  \\
\bottomrule
\end{tabular}
\end{minipage}
\hfill  
\begin{minipage}{0.43\textwidth}
\centering
\caption{Ablation on the number of bins $B$ in the angular distribution.}
\vspace{-3mm}
\renewcommand{\arraystretch}{0.9}
\label{tab:ablation_2d_bin}
\setlength{\tabcolsep}{2.2pt}
\begin{tabular}{l|cc>{\columncolor{black!5}}ccc}
\toprule
TT-Barn & $12$ & $24$ &  $36$ & $48$ & $72$ \\
\midrule
NUM & 0.903 & 0.911 & 0.903 &  0.938 & 0.962   \\
GPU & 6571 & 6827 & 6569 & 6901 & 7123 \\
PSNR & 25.63 & 25.71 &  25.90 & 25.81 & 25.88  \\
\bottomrule
\end{tabular}
\end{minipage}
\vspace{-4mm}
\end{table}

\vspace{1mm}
\noindent\textbf{Effect of 3D Frequency Filter.}
We analyze 3D frequency filter by varying the reference frequency $t$ and exponent $\alpha$ based on NS-chair. As summarized in \Cref{tab:ablation_3d_filter}, we investigate the reference frequency from $4$ to $10$ with the exponent within $1$, $2$, $3$, and $4$. The results remain stable across a range of settings, and the best result is achieved at $t$ as $8$ and $\alpha$ as $2$. We also examine the pruning ratio $\rho$ and view sampling number $K^\ast$ in \Cref{tab:ablation_3d_prune}. The results reveal that a higher $\rho$ aggressively suppresses Gaussian growth, and the best trade-off is achieved at $\rho$ as $3\%$ with $K^\ast$ as $48$, which consistently delivers strong SSIM and PSNR while comparatively keeping reasonable Gaussian number and memory cost. This confirms that moderate frequency filtering effectively removes redundant high-frequency Gaussians to preserve reconstruction accuracy.

\vspace{1mm}
\noindent\textbf{Effect of 2D Spectral Regularization.}
We ablate 2D spectral regularization by evaluating frequency thresholds $\dot \gamma_{\text{min}}$ and $\dot \gamma_{\text{max}}$ and the number of angular bins $B$ using the scene from TT-Barn. As shown in \Cref{tab:ablation_2d_gamma} and \Cref{tab:ablation_2d_bin} across all settings, the differences in Gaussian number, GPU memory, and PSNR remain small, indicating that the method is largely insensitive to its hyper-parameter settings. For the frequency thresholds, values between $0.3$ and $0.9$ consistently yield slightly better fidelity while keeping a compact model size. For the anisotropy control, we constrain directional high-frequency structures by using the number of angular bins $B$ as $36$, where a larger $\dot \gamma_{\text{max}}$ includes more high-frequency energy, and a larger $B$ sharpens the angular histogram. Overall, these results confirm that the 2D spectral regularization is stable and robust, with weak dependency on the specific hyper-parameter choices.

\begin{table}[t]
\scriptsize
\centering
\begin{minipage}{0.53\textwidth}
\centering
\caption{Ablation under different attack strengths. We report poison/defense results.}
\vspace{-3mm}
\label{tab:exp_attack_range}
\renewcommand{\arraystretch}{0.9}
\setlength{\tabcolsep}{3.2pt}
\begin{tabular}{l|l|c|c}
\toprule
Attack $\epsilon$ & Metric & NUM & GPU \\
\midrule
\multirow{3}{*}{\makecell[l]{Strength: \\ $8/255$}} & NS-chair  & 0.672/0.284 & 5094/3021  \\
& NS-drums  & 0.495/0.210 & 4739/3006  \\
& NS-lego & 0.485/0.271 & 4664/3108   \\
\midrule
\multirow{3}{*}{\makecell[l]{Strength: \\ $24/255$}} & NS-chair & 1.158/0.665  & 14223/6340 \\
& NS-drums & 0.839/0.367 & 12658/5907 \\
& NS-lego & 1.107/0.496 & 16296/7525 \\
\midrule
\multirow{3}{*}{\makecell[l]{Strength: \\ $\infty$}} & NS-chair & 4.223/1.855 & 47425/28518  \\
& NS-drums & 3.998/1.936 & 47386/30723  \\
& NS-lego & 4.100/1.921 & 47811/32090 \\
\bottomrule
\end{tabular}
\end{minipage}
\hfill
\begin{minipage}{0.44\textwidth}
\centering
\caption{Ablation on loss weight to balance the overall training loss.}
\vspace{-3mm}
\label{tab:ablation_loss_weights}
\renewcommand{\arraystretch}{0.9}
\setlength{\tabcolsep}{1.3pt}
\begin{tabular}{l|cc|cc|cc}
\toprule
\multirow{2}{*}{$\lambda$} & \multicolumn{2}{c|}{TT-Barn} & \multicolumn{2}{c|}{NS-chair} & \multicolumn{2}{c}{MIP-bonsai} \\
& NUM  & PSNR & NUM  & PSNR & NUM  & PSNR \\
\midrule
0   & 1.553 & 25.17 & 0.477 & 24.73 & 1.712 & 27.59 \\
\midrule
1   & 1.416 & 25.12 & 0.421 & 24.86 & 1.454 & 28.01 \\
\midrule
2   & 1.349 & 25.03 & 0.395 & 24.52 & 1.390 & 29.02 \\
\midrule
3   & 1.121 & 25.62 &  0.329 & 24.91 & 1.228 & 28.65 \\ 
\midrule
4   & \cellcolor{black!5} 0.903 & \cellcolor{black!5} 25.90 & \cellcolor{black!5} 0.281 & \cellcolor{black!5} 25.00 & 1.106 & 28.52  \\
\midrule
5   & 0.877 & 25.54 & 0.236 & 24.31 & \cellcolor{black!5} 1.037 & \cellcolor{black!5} 29.07 \\
\bottomrule
\end{tabular}
\end{minipage}
\vspace{-3mm}
\end{table}

\vspace{1mm}
\noindent\textbf{Effect of Various Attack Strengths.}
In addition to default attack strength with $\epsilon$ as $16/255$, we report poison/defense results on constrained attack $8/255$ and $24/255$, and unconstrained attack $\infty$. \Cref{tab:exp_attack_range} provides various perturbations under different attack strengths, demonstrating our defense effects.

\vspace{1mm}
\noindent\textbf{Effect of Loss Weight.}
\Cref{tab:ablation_loss_weights} analyzes the influence of loss weight $\lambda$ on three representative scenes from TT-Barn, NS-chair, and MIP-bonsai. Increasing $\lambda$ consistently reduces the number of Gaussians by suppressing poisoned overgrowth, while slightly improving PSNR. For the MIP-bonsai, a large loss weight continues to prune Gaussians effectively, and $\lambda$ as $5$ provides a strong suppression without sacrificing fidelity. However, a large $\lambda$ over-suppresses natural patterns on small scene details, causing a slight PSNR drop, particularly on the NS-chair. The intermediate setting for NS-chair is $\lambda$ as $4$, providing the most balanced performance. 

\vspace{-2mm}
\section{Conclusion}
\label{sec:con}
\vspace{-2mm}
We address the defense approach to the resource-targeting attack that triggers poisoned Gaussian overgrowth in 3DGS. By proposing a spectral defense, our 3D frequency filter and 2D spectral regularization realize selective pruning and anisotropic suppression, establishing the first robust defense framework for secure and efficient 3DGS.

\par\vfill\par
\clearpage  


%
%
\bibliographystyle{splncs04}
\bibliography{main}

@String(PAMI  = {IEEE Trans. Pattern Anal. Mach. Intell.})

@String(CVPR  = {IEEE Conf. Comput. Vis. Pattern Recog.})

@String(ICCV  = {Int. Conf. Comput. Vis.})

@String(ECCV  = {Eur. Conf. Comput. Vis.})

@String(NeurIPS = {Adv. Neural Inform. Process. Syst.})

@String(ICLR  = {Int. Conf. Learn. Represent.})

@String(AAAI  = {AAAI Conf. on Artificial Intelligence})

@String(TOG   = {ACM Trans. Graph.})

@String(TCSVT = {IEEE Trans. Circuit Syst. Video Technol.})

@String(ACMMM = {ACM Int. Conf. Multimedia})

@STRING{TIFS        = "{IEEE} Trans. on Information Forensics and Security"}

@article{knapitsch2017tanks,
  title={Tanks and temples: Benchmarking large-scale scene reconstruction},
  author={Knapitsch, Arno and Park, Jaesik and Zhou, Qian-Yi and Koltun, Vladlen},
  journal=ToG,
  volume={36},
  number={4},
  pages={1--13},
  year={2017}
}

@article{mildenhall2021nerf,
  title={Nerf: Representing scenes as neural radiance fields for view synthesis},
  author={Mildenhall, Ben and Srinivasan, Pratul P and Tancik, Matthew and Barron, Jonathan T and Ramamoorthi, Ravi and Ng, Ren},
  journal=CACM,
  volume={65},
  number={1},
  pages={99--106},
  year={2021}
}

@inproceedings{barron2022mip,
  title={Mip-nerf 360: Unbounded anti-aliased neural radiance fields},
  author={Barron, Jonathan T and Mildenhall, Ben and Verbin, Dor and Srinivasan, Pratul P and Hedman, Peter},
  booktitle=CVPR,
  pages={5470--5479},
  year={2022}
}

@article{kerbl20233d,
  title={3d gaussian splatting for real-time radiance field rendering.},
  author={Kerbl, Bernhard and Kopanas, Georgios and Leimk{\"u}hler, Thomas and Drettakis, George},
  journal=ToG,
  volume={42},
  number={4},
  pages={139--1},
  year={2023}
}

@inproceedings{schonberger2016structure,
  title={Structure-from-motion revisited},
  author={Schonberger, Johannes L and Frahm, Jan-Michael},
  booktitle=CVPR,
  pages={4104--4113},
  year={2016}
}

@inproceedings{lu2024poison,
  title={Poison-splat: Computation cost attack on 3d gaussian splatting},
  author={Lu, Jiahao and Zhang, Yifan and Shen, Qiuhong and Wang, Xinchao and Yan, Shuicheng},
  booktitle=ICLR,
  year={2025}
}

@inproceedings{fan2024lightgaussian,
  title={Lightgaussian: Unbounded 3d gaussian compression with 15x reduction and 200+ fps},
  author={Fan, Zhiwen and Wang, Kevin and Wen, Kairun and Zhu, Zehao and Xu, Dejia and Wang, Zhangyang and others},
  booktitle=NeurIPS,
  volume={37},
  pages={140138--140158},
  year={2024}
}

@inproceedings{hanson2025pup,
  title={Pup 3d-gs: Principled uncertainty pruning for 3d gaussian splatting},
  author={Hanson, Alex and Tu, Allen and Singla, Vasu and Jayawardhana, Mayuka and Zwicker, Matthias and Goldstein, Tom},
  booktitle=CVPR,
  pages={5949--5958},
  year={2025}
}

@article{bao20253d,
  title={3d gaussian splatting: Survey, technologies, challenges, and opportunities},
  author={Bao, Yanqi and Ding, Tianyu and Huo, Jing and Liu, Yaoli and Li, Yuxin and Li, Wenbin and Gao, Yang and Luo, Jiebo},
  journal=TCSVT,
  year={2025}
}

@inproceedings{xu2025depthsplat,
  title={Depthsplat: Connecting gaussian splatting and depth},
  author={Xu, Haofei and Peng, Songyou and Wang, Fangjinhua and Blum, Hermann and Barath, Daniel and Geiger, Andreas and Pollefeys, Marc},
  booktitle=CVPR,
  pages={16453--16463},
  year={2025}
}

@article{lei2025gaussnav,
  title={Gaussnav: Gaussian splatting for visual navigation},
  author={Lei, Xiaohan and Wang, Min and Zhou, Wengang and Li, Houqiang},
  journal=PAMI,
  year={2025},
  publisher={IEEE}
}

@inproceedings{kong2025rogsplat,
  title={Rogsplat: Robust gaussian splatting via generative priors},
  author={Kong, Hanyang and Yang, Xingyi and Wang, Xinchao},
  booktitle=ICCV,
  pages={25735--25745},
  year={2025}
}

@inproceedings{zhang2025quadratic,
  title={Quadratic Gaussian Splatting: High Quality Surface Reconstruction with Second-order Geometric Primitives},
  author={Zhang, Ziyu and Huang, Binbin and Jiang, Hanqing and Zhou, Liyang and Xiang, Xiaojun and Shen, Shuhan},
  booktitle=ICCV,
  pages={28260--28270},
  year={2025}
}

@inproceedings{shumailov2021sponge,
  title={Sponge examples: Energy-latency attacks on neural networks},
  author={Shumailov, Ilia and Zhao, Yiren and Bates, Daniel and Papernot, Nicolas and Mullins, Robert and Anderson, Ross},
  booktitle={IEEE EuroS\&P},
  pages={212--231},
  year={2021},
}

@inproceedings{chen2023dark,
  title={The dark side of dynamic routing neural networks: Towards efficiency backdoor injection},
  author={Chen, Simin and Chen, Hanlin and Haque, Mirazul and Liu, Cong and Yang, Wei},
  booktitle=CVPR,
  pages={24585--24594},
  year={2023}
}

@article{cina2025energy,
  title={Energy-latency attacks via sponge poisoning},
  author={Cin{\`a}, Antonio Emanuele and Demontis, Ambra and Biggio, Battista and Roli, Fabio and Pelillo, Marcello},
  journal={Information Sciences},
  volume={702},
  pages={121905},
  year={2025}
}

@inproceedings{tang2024lgm,
  title={Lgm: Large multi-view gaussian model for high-resolution 3d content creation},
  author={Tang, Jiaxiang and Chen, Zhaoxi and Chen, Xiaokang and Wang, Tengfei and Zeng, Gang and Liu, Ziwei},
  booktitle=ECCV,
  pages={1--18},
  year={2024}
}

@inproceedings{chen2024text,
  title={Text-to-3d using gaussian splatting},
  author={Chen, Zilong and Wang, Feng and Wang, Yikai and Liu, Huaping},
  booktitle=CVPR,
  pages={21401--21412},
  year={2024}
}

@inproceedings{wen2025segs,
  title={Segs-slam: Structure-enhanced 3d gaussian splatting slam with appearance embedding},
  author={Wen, Tianci and Liu, Zhiang and Fang, Yongchun},
  booktitle=ICCV,
  pages={28103--28113},
  year={2025}
}

@inproceedings{homeyer2025droid,
  title={DROID-Splat Combining end-to-end SLAM with 3D Gaussian Splatting},
  author={Homeyer, Christian and Begiristain, Leon and Schn{\"o}rr, Christoph},
  booktitle=ICCV,
  pages={2767--2777},
  year={2025}
}

@inproceedings{xu2025sequential,
  title={Sequential Gaussian Avatars with Hierarchical Motion Context},
  author={Xu, Wangze and Zhan, Yifan and Zhong, Zhihang and Sun, Xiao},
  booktitle=ICCV,
  pages={13592--13603},
  year={2025}
}

@inproceedings{li20253d,
  title={3D Gaussian Representations with Motion Trajectory Field for Monocular Dynamic Scene Reconstruction},
  author={Li, Xuesong and Petersson, Lars and Rolland, Vivien},
  booktitle=ICCV,
  pages={6430--6439},
  year={2025}
}

@inproceedings{niedermayr2024compressed,
  title={Compressed 3d gaussian splatting for accelerated novel view synthesis},
  author={Niedermayr, Simon and Stumpfegger, Josef and Westermann, R{\"u}diger},
  booktitle=CVPR,
  pages={10349--10358},
  year={2024}
}

@inproceedings{lee2024compact,
  title={Compact 3d gaussian representation for radiance field},
  author={Lee, Joo Chan and Rho, Daniel and Sun, Xiangyu and Ko, Jong Hwan and Park, Eunbyung},
  booktitle=CVPR,
  pages={21719--21728},
  year={2024}
}

@inproceedings{chen2024hac,
  title={Hac: Hash-grid assisted context for 3d gaussian splatting compression},
  author={Chen, Yihang and Wu, Qianyi and Lin, Weiyao and Harandi, Mehrtash and Cai, Jianfei},
  booktitle={European Conference on Computer Vision},
  pages={422--438},
  year={2024},
  organization={Springer}
}

@inproceedings{liu2024compgs,
  title={Compgs: Efficient 3d scene representation via compressed gaussian splatting},
  author={Liu, Xiangrui and Wu, Xinju and Zhang, Pingping and Wang, Shiqi and Li, Zhu and Kwong, Sam},
  booktitle=ACMMM,
  pages={2936--2944},
  year={2024}
}

@inproceedings{zhang2024lp,
  title={Lp-3dgs: Learning to prune 3d gaussian splatting},
  author={Zhang, Zhaoliang and Song, Tianchen and Lee, Yongjae and Yang, Li and Peng, Cheng and Chellappa, Rama and Fan, Deliang},
  booktitle=NeurIPS,
  pages={122434--122457},
  year={2024}
}

@inproceedings{liu2025maskgaussian,
  title={Maskgaussian: Adaptive 3d gaussian representation from probabilistic masks},
  author={Liu, Yifei and Zhong, Zhihang and Zhan, Yifan and Xu, Sheng and Sun, Xiao},
  booktitle=CVPR,
  pages={681--690},
  year={2025}
}

@inproceedings{fang2024mini,
  title={Mini-splatting: Representing scenes with a constrained number of gaussians},
  author={Fang, Guangchi and Wang, Bing},
  booktitle=ECCV,
  pages={165--181},
  year={2024}
}

@inproceedings{horvath2023targeted,
  title={Targeted adversarial attacks on generalizable neural radiance fields},
  author={Horv{\'a}th, Andr{\'a}s and J{\'o}zsa, Csaba M},
  booktitle=ICCV,
  pages={3718--3727},
  year={2023}
}

@inproceedings{jiang2024nerfail,
  title={Nerfail: Neural radiance fields-based multiview adversarial attack},
  author={Jiang, Wenxiang and Zhang, Hanwei and Wang, Xi and Guo, Zhongwen and Wang, Hao},
  booktitle=AAAI,
  pages={21197--21205},
  year={2024}
}

@inproceedings{huang2024towards,
  title={Towards transferable targeted 3d adversarial attack in the physical world},
  author={Huang, Yao and Dong, Yinpeng and Ruan, Shouwei and Yang, Xiao and Su, Hang and Wei, Xingxing},
  booktitle=CVPR,
  pages={24512--24522},
  year={2024}
}

@article{jiang2024ipa,
  title={Ipa-nerf: Illusory poisoning attack against neural radiance fields},
  author={Jiang, Wenxiang and Zhang, Hanwei and Zhao, Shuo and Guo, Zhongwen and Wang, Hao},
  journal={arXiv preprint arXiv:2407.11921},
  year={2024}
}

@article{zeybey2024gaussian,
  title={Gaussian Splatting Under Attack: Investigating Adversarial Noise in 3D Objects},
  author={Zeybey, Abdurrahman and Ergezer, Mehmet and Nguyen, Tommy},
  journal={arXiv preprint arXiv:2412.02803},
  year={2024}
}

@inproceedings{ke2025stealthattack,
  title={StealthAttack: Robust 3D Gaussian Splatting Poisoning via Density-Guided Illusions},
  author={Ke, Bo-Hsu and Xie, You-Zhe and Liu, Yu-Lun and Chiu, Wei-Chen},
  booktitle=ICCV,
  pages={27400--27411},
  year={2025}
}

@inproceedings{ruan2023towards,
  title={Towards viewpoint-invariant visual recognition via adversarial training},
  author={Ruan, Shouwei and Dong, Yinpeng and Su, Hang and Peng, Jianteng and Chen, Ning and Wei, Xingxing},
  booktitle=ICCV,
  pages={4709--4719},
  year={2023}
}

@article{kuang2024defense,
  title={Defense against adversarial attacks using topology aligning adversarial training},
  author={Kuang, Huafeng and Liu, Hong and Lin, Xianming and Ji, Rongrong},
  journal=TIFS,
  volume={19},
  pages={3659--3673},
  year={2024}
}

@inproceedings{madry2018towards,
  title={Towards deep learning models resistant to adversarial attacks},
  author={Madry, Aleksander and Makelov, Aleksandar and Schmidt, Ludwig and Tsipras, Dimitris and Vladu, Adrian},
  booktitle=ICLR,
  year={2018}
}

@inproceedings{lu2024scaffold,
  title={Scaffold-gs: Structured 3d gaussians for view-adaptive rendering},
  author={Lu, Tao and Yu, Mulin and Xu, Linning and Xiangli, Yuanbo and Wang, Limin and Lin, Dahua and Dai, Bo},
  booktitle=CVPR,
  pages={20654--20664},
  year={2024}
}

\clearpage

\end{document}